\pdfoutput=1

\documentclass[11pt]{article}

\usepackage[final]{acl}

\usepackage{times}
\usepackage{latexsym}
\usepackage{csquotes}
\usepackage{booktabs}
\usepackage{multirow} 
\usepackage{array}
\newcolumntype{M}[1]{>{\centering\arraybackslash}m{#1}}
\usepackage[T1]{fontenc}

\usepackage[utf8]{inputenc}

\usepackage{microtype}

\usepackage{inconsolata}

\usepackage{graphicx}
\usepackage{algorithm}
\usepackage{algpseudocode}
\usepackage{amsmath}

\usepackage[nolist]{acronym}
\begin{acronym}
    \acro{HCI}{Human-Computer Interaction}
    \acro{LLM}{Large Language Model}
    \acro{CS}{Computer Science}
    \acro{RAG}{Retrieval-Augmented Generation}
    \acro{AI}{Artificial Intelligent}
    \acro{MMR}{Maximal Marginal Relevance}
    \acro{VL}{Visual Language}
    \acro{MWP}{math word problem}
     \acro{MWPs}{math word problems}
     \acro{TTI}{Text-to-Image}
     \acro{MCQs}{multiple-choice questions}
\end{acronym}
\usepackage{listings}
\usepackage{xcolor}
\usepackage[utf8]{inputenc}  

\usepackage{tikz}
\newcommand*\circled[1]{\tikz[baseline=(char.base)]{
            \node[shape=circle,draw,inner sep=1pt] (char) {#1};}}
\usepackage{todonotes}
\usepackage{marginnote}

\definecolor{TodoColor}{rgb}{1,0.7,0.6}

\newcommand{\sys}{\textsc{Math2Visual}}
\newcommand{\mw}{MathWorld}

\newcommand{\img}{\protect\includegraphics[height=1em]{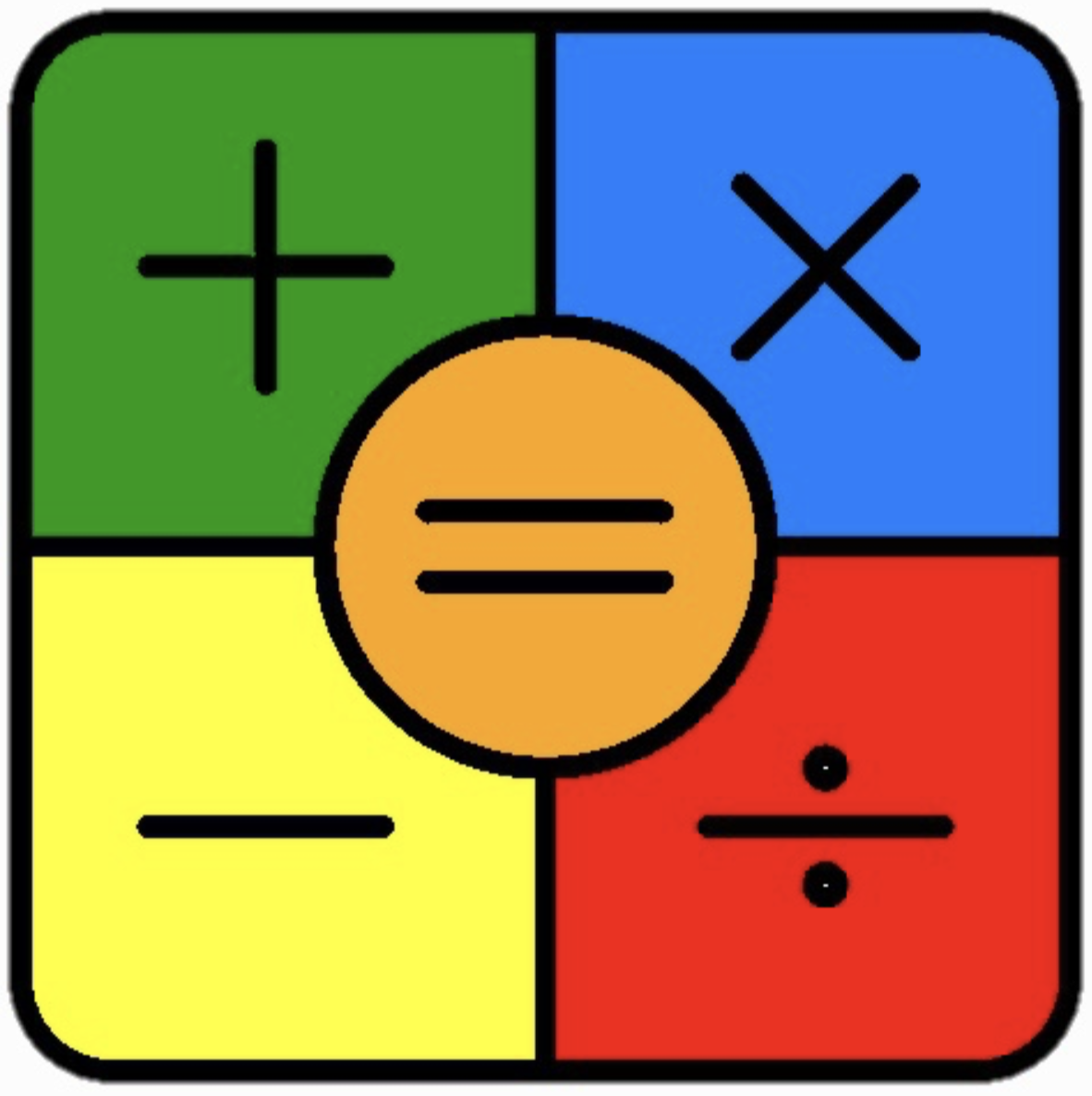}~}

\definecolor{FindingsColor}{gray}{0.85}

\makeatletter\def\Hy@Warning#1{}\makeatother
\let\svthefootnote\thefootnote
\newcommand\blankfootnote[1]{%
  \let\thefootnote\relax\footnotetext{#1}%
  \let\thefootnote\svthefootnote%
}

\title{\img Generating Pedagogically Meaningful Visuals for Math Word Problems: A New Benchmark and Analysis of Text-to-Image Models}

\author{
    Junling Wang$^{1, 2}$ \quad
    Anna Rutkiewicz$^{3}$ \quad
    April Yi Wang$^{1}$ \quad
    \textbf{
    Mrinmaya Sachan$^{1}$
    } \\ \text{} \\
  $^{1}$Department of Computer Science, ETH Zurich \quad \\
  $^2$ ETH AI Center \\
  $^{3}$ Department of Informatics, University of Zurich \\
}

\begin{document}
\maketitle

\begin{abstract}
Visuals are valuable tools for teaching \ac{MWPs}, helping young learners interpret textual descriptions into mathematical expressions before solving them.
However, creating such visuals is labor-intensive and there is a lack of automated methods to support this process. In this paper, we present \sys{}, an automatic framework for 
generating  pedagogically meaningful visuals from MWP text descriptions. 
\sys{} leverages a pre-defined visual language and a design space grounded in interviews with math teachers, to illustrate the core mathematical relationships in \ac{MWPs}.
Using \sys{}, we construct an annotated dataset of 1,903 visuals and evaluate \ac{TTI} models for their ability to generate visuals that align with our design. 
We further fine-tune several \ac{TTI} models with our dataset, demonstrating improvements in educational visual generation. 
Our work establishes a new benchmark for automated generation of pedagogically meaningful visuals and offers insights into key challenges in producing multimodal educational content, such as the misrepresentation of mathematical relationships and the omission of essential visual elements.

\hspace{.5em}\includegraphics[width=1.25em,height=1.25em]{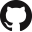}\hspace{.75em}\parbox{\dimexpr\linewidth-2\fboxsep-2\fboxrule}{\url{https://github.com/eth-lre/math2visual}}

\end{abstract}
\section{Introduction}

Math word problems (MWPs) describe mathematical scenarios through text, requiring learners to interpret both linguistic and numerical information to derive mathematical expressions for problem-solving~\citep{Verschaffel2014}. MWPs are a key component of primary school math education and have been the subject of significant educational research~\citep{verschaffel2020word}.
Solving MWPs is a complex cognitive task that 
progresses through several stages: problem understanding, solution planning and solution execution~\citep{opedal-etal-2023-world,polya2014solve}.
A major challenge lies in interpreting the text and constructing a mental model that captures the underlying mathematical relationships~\cite{CUMMINS1988405, stern1993makes} --- a process especially difficult for young students (e.g., Grades 1–3) who are still developing their reading and comprehension skills~\citep{duke2012improving}.
Moreover, recent findings 
suggest that children’s arithmetic skills do not readily transfer between applied and academic contexts~\citep{banerjee2025children}, highlighting the need to bridge everyday experiences with formal instruction.
Visual representations designed specifically for MWPs offer a promising solution: by translating textual descriptions into intuitive forms~\citep{cooper2018benefits}, they help learners map language to mathematical structure, thereby supporting comprehension and problem solving~\citep{MAYER200285}.

\begin{figure}
    \centering
    \includegraphics[width=0.3\textwidth]{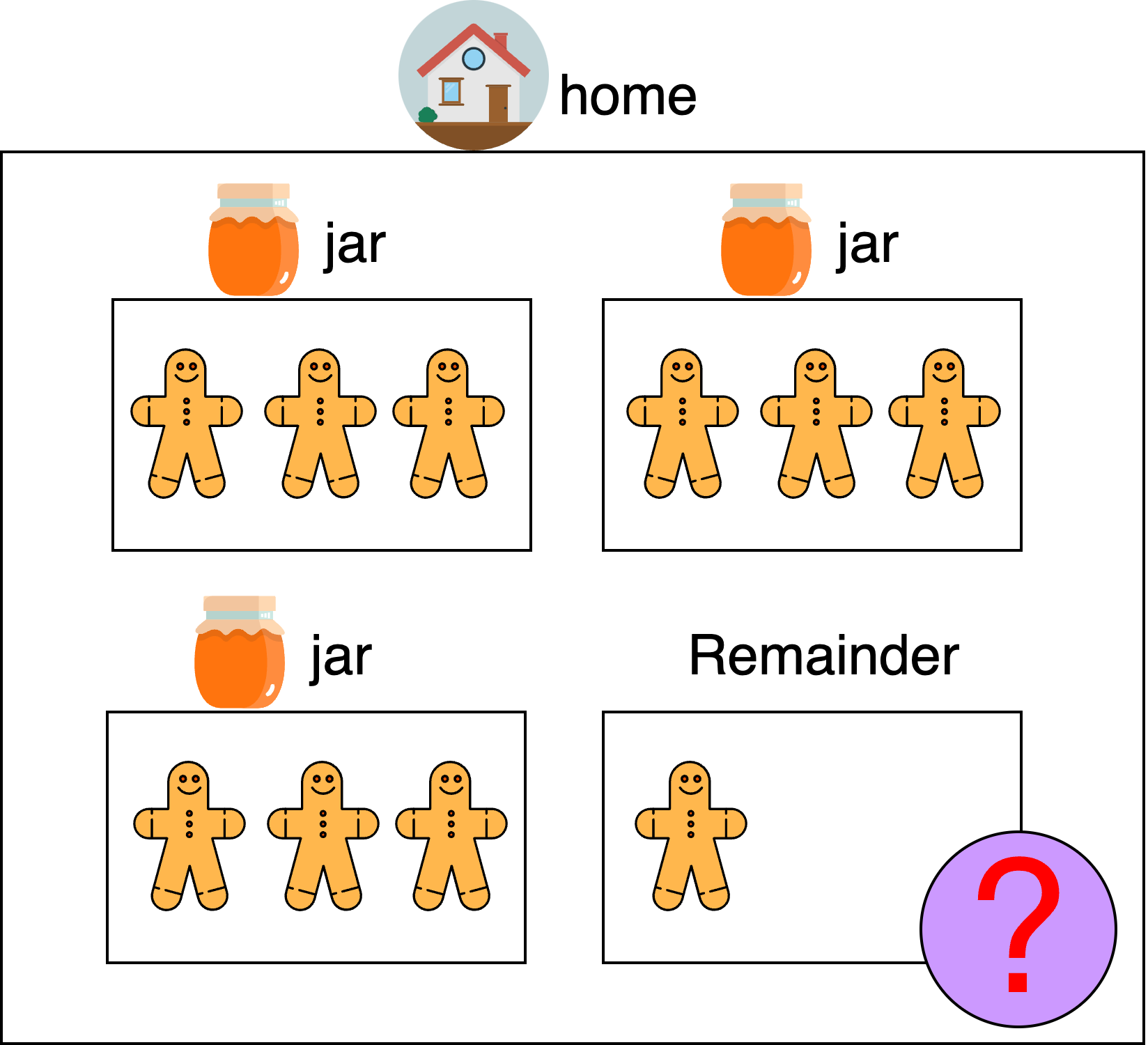}
    \caption{Surplus operation example in Intuitive design (Formal version: Figure~\ref{fig:fig19}). MWP: At home, Marian made 10 gingerbread cookies, which she will distribute equally among tiny glass jars. If each jar is to contain 3 cookies, how many cookies will remain unplaced?}
    \label{fig:fig0}
    \vspace{-15pt}
\end{figure}
Although primary school math teachers have long recognized the value of visuals when teaching MWPs~\citep{kaitera2022developing,boonen2016s}, manually creating these visuals is time-consuming and requires considerable effort~\citep{xu2021procedural}.
Recent advances in \acf{TTI} models offer potential for automating visual generation, but current models often fail to capture the underlying mathematical reasoning required for MWPs~\citep{kajic2024evaluating}.
In response, prior work has explored automating instructional image generation and retrieval.
For instance, Singh et al. introduced a text-image matching task aimed at retrieving and assigning web images to textbook content~\citep{singh-etal-2023-enhancing}, and later explored using image semantics to generate visual multiple-choice questions for early learners~\citep{singh2019automatic}. 
However, these methods are not designed to handle narrative-driven problems like MWPs, which require grounding visual content in contextualized scenarios. 
VisualMath made early attempts to visualize MWPs using existing images, but covers only basic operations and provides no discussion of the pedagogical grounding or validation of its visual design through collaboration with educators~\citep{dwivedi2017visualmath}.
To date, there is no established framework for generating visuals that are both pedagogically meaningful and scalable for diverse narrative structures found in MWPs.

In response to these gaps, we co-design a pedagogically meaningful visual design for MWPs with primary school math teachers. 
Here, we define pedagogically meaningful visuals as those that semantically and logically represent the mathematical structure of a word problem, thereby helping learners in accurately and clearly comprehending its content.
Then, we introduce \sys{}, a framework for generating such visuals from MWP text descriptions. 
Using \sys{}, we generate and annotate a dataset containing $\sim$2K pedagogical visuals for MWPs in Grades 1–3.
Finally, we evaluate the ability of state-of-the-art \ac{TTI} models to directly generate visuals aligned with our proposed pedagogical design. By fine-tuning these models on our annotated dataset, we demonstrate notable improvements in generation quality. In summary, our contributions are:


\noindent\circled{1} \sys{}, a scalable framework that incorporates a tree-based visual language and a structured design space to generate pedagogically meaningful visuals from MWP text descriptions.

\noindent\circled{2} An annotated visual dataset that benchmarks models' ability to generate mathematically reasoned visuals and supports \ac{TTI} model training.

\section{Related Work}
\paragraph{Math Word Problems in NLP}
Math word problems have long been a focus of interest in the NLP community~\citep{roy-roth-2015-solving,kushman-etal-2014-learning,huang-etal-2017-learning,amini-etal-2019-mathqa,ijcai2019p736,drori2002}, with research primarily aiming to improve computational models’ ability to solve MWPs accurately. Approaches such as mapping text to expression trees~\citep{koncel2015,yang-etal-2022-logicsolver,roy2017} and explicitly modeling arithmetic operations~\citep{mitra-baral-2016-learning,roy2018} have enhanced machine processing of mathematical expressions in natural language. However, most existing methods focus on producing numerical answers without human-interpretable reasoning, which is essential in educational settings~\citep{opedal-etal-2023-world,shridhar-etal-2022-automatic}. To address this limitation, recent work has explored integrating mental models and human-centered representations into MWP solving. The \mw{} framework~\citep{opedal-etal-2023-world} represents MWPs using a graph-based semantic formalism aligned with human reasoning. However, it supports only the four basic arithmetic operations, and lacks coverage of ``second-order'' MWPs.

\paragraph{Visuals in Primary School Math Education}\label{sec:literature}
Visuals have long been recognized as critical tools in primary school education, particularly in math teaching~\citep{kaitera2022developing,boonen2016s}. Research indicates that well-designed pedagogical visuals help students grasp abstract concepts more readily~\citep{small2025eyes,MAYER200285,evagorou2015role} while increasing their engagement~\citep{cooper2018benefits}, and improving study efficiency~\citep{arcavi2003role}. Many visual designs have been proposed for primary school math teaching. One common design is bar model~\cite{hoven2007singapore}.
The bar model illustrates numerical relationships of math problems through bars representing quantities, enabling visualization of mathematical concepts and operations~\citep{hoven2007singapore}. 
The bar model has proven to be effective in improving children's problem solving skills~\citep{Osman2018} and their ability to use correct cognitive strategies to solve the problem~\citep{morin2017use}.
Another modern design is the Noyon framework, which introduces a modular approach to visually expressing mathematical problems~\citep{saquib2021constructing}. 
Noyon employs iconic elements to construct representations of mathematical concepts, offering a structured yet flexible way to depict mathematical relationships.


\paragraph{Automated Visual Generation and Retrieval in Education}
Although educational visuals are widely recognized for their benefits and are frequently used by primary school math teachers in instruction~\citep{JITENDRA2019269,boonen2016s}, the manual creation of such visuals remains a time-consuming and resource-intensive task~\citep{xu2021procedural}.
Recent advances in NLP and educational technology have explored automated methods for generating or retrieving visual content. For instance, tasks such as text-image matching have been proposed to assign web images to textbook content~\citep{singh-etal-2023-enhancing}, while other studies have leveraged image semantics to generate visual multiple-choice questions~\citep{singh2019automatic} and employed frameworks like Chain-of-Exemplar to combine multimodal educational content for question generation~\citep{luo-etal-2024-chain}. However, these approaches fail to generate visuals that reveal the underlying mathematical reasoning in MWPs. The VisualMath proposed a system for visualizing MWPs using existing images but covers only basic operations (+, -) and provides no discussion of the pedagogical grounding or validation of its visual design through collaboration with educators~\citep{dwivedi2017visualmath}.

\begin{figure*}[h!]
    \centering
    \includegraphics[width=\textwidth]{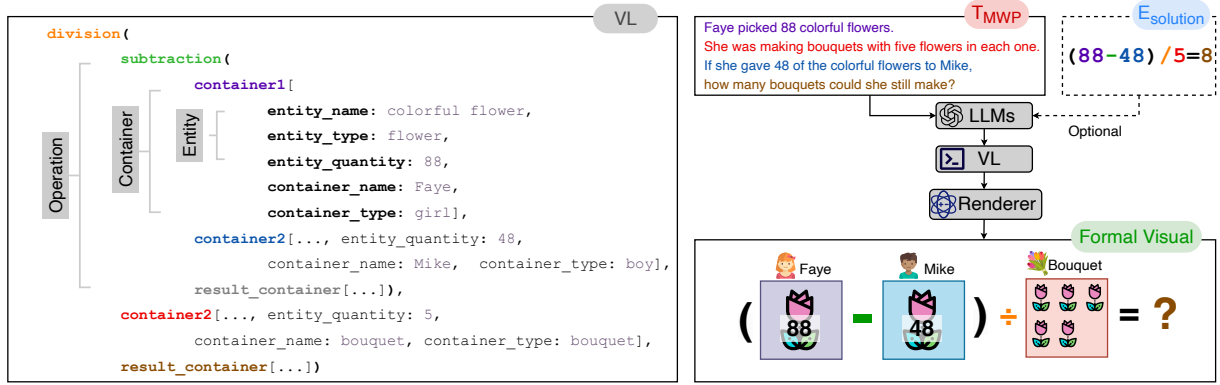}
    \caption{\sys{} Framework: Our approach first converts the MWP text description into a \acf{VL} expression using an LLM. The \ac{VL} is then passed to a rendering program that generates the corresponding visual. The presented visual is in ``Formal'' design.
    }
    \label{fig:fig2}
    \vspace{-15pt}
\end{figure*}


\section{From MWP to Visual}
This section introduces \sys{} framework. We first present the desiderata for a good visual (Section~\ref{sec:design_goal}), followed by an overview of \sys{} ( Section~\ref{sec:math2visual_overview}). Then, we explain each component of \sys{} (Section~\ref{sec:visual_language} to \ref{sec:render}). Finally, we detail the process of developing visual designs with teachers and the evaluation criteria (Sections~\ref{sec:design_expert} and \ref{sec:eval_criteria}).

\subsection{Desiderata for a Good Visual}\label{sec:design_goal}
For visuals aimed at supporting primary school educators and enhancing student understanding of MWPs (Grades 1–3), the following criteria are essential:
(1) clearly convey the central ideas of an MWP~\citep{evagorou2015role,JITENDRA2019269}, (2) prevent unnecessary cognitive load of students~\citep{MAYER200285}, and (3) enhance student engagement~\citep{cooper2018benefits}. 
Rather than focusing on decorative aesthetics, the design should maintain a semantic and logical alignment with the MWP content~\citep{sahinkaya2024visualizing}. 
\subsection{\sys{} Framework Overview}\label{sec:math2visual_overview}
Our work focuses on simple MWPs --- problems where a single mathematical expression leads to a solution, which are common in early math education (Grades 1-3). In this context, we introduce the \sys{} framework for generating two specific types of educational visuals: 

\noindent(1) \textbf{Formal visuals}, which depict mathematical relationships in a symbolic style. These visuals are designed to help learners understand the underlying math relationships in a clear and mathematical way. 

\noindent(2) \textbf{Intuitive visuals}, which represent mathematical relationships in a context-rich, example-based way that mimics real-world scenarios or story settings. 
These visuals are designed to improve engagement and reduce the cognitive load of students. More details about the visual design process are shown in Section~\ref{sec:design_expert}.

An overview of the \sys{} pipeline is shown in Figure~\ref{fig:fig2}. 
\sys{}  follows a text-to-semantics-to-visual pipeline, similar to previous visual generation works~\citep{belouadi2023automatikz,belouadi2024detikzify}. Given an MWP text description ($T_{\text{MWP}}$) and, optionally, a solution expression (\(E_{\text{solution}}\)), the framework uses an LLM to produce a visual language \ac{VL}. \ac{VL} is a semantic visual representation that holds the information needed to generate the visuals (see Section~\ref{sec:visual_language}). The \ac{VL} is then paired with a manually collected dataset of icons, called \(SVG\) and processed by two rendering programs (\(R_{\text{formal}}\), \(R_{\text{intuitive}}\)) to generate two types of visuals: ``Formal'' (\(V_{\text{formal}}\)) and ``Intuitive'' (\(V_{\text{intuitive}}\)). Details of the visual design and rendering program are presented in Sections~\ref{sec:design_element} and~\ref{sec:render}, respectively.

\subsection{Semantic Representation of MWP}\label{sec:visual_language}

To bridge the gap between formal mathematical structure and visual expressiveness, we introduce a tree-structured 
\acf{VL} specifically tailored for visual generation. 

\ac{VL} is a 
hierarchical language with a structure closely resembling the
expression tree ~\citep{wang-etal-2018-translating,zhang-etal-2023-expression} of the solution expression \(E_{\text{solution}}\). 
In the \ac{VL}, we represent an MWP using three primary components: entity, container, and operation. We illustrate the mapping from an MWP to these \ac{VL} components using the example in Figure~\ref{fig:fig2}. Note that the mapping from an MWP to \ac{VL} is not strictly deterministic --- it requires an intuitive understanding of the visualization. 
Therefore, we use an LLM with in-context learning to perform the conversion. The full procedure is detailed in Section~\ref{sec:dataset_creation}.
 
\noindent(1) \textbf{Entity} is the smallest unit in \ac{VL} and represents an element to be visualized. 
For instance, the flower in Figure~\ref{fig:fig2} is an entity.
An entity is identified by attributes \texttt{entity\_name},  \texttt{entity\_type} and \texttt{entity\_quantity}. 
The \texttt{entity\_name} represents the name of the entity as given in the MWP, \texttt{entity\_type} is entity's category for visualization. 
In Figure~\ref{fig:fig2}, the phrase ``colorful flower'' from the MWP maps to \texttt{entity\_name}, while ``flower'' becomes the \texttt{entity\_type}. The \texttt{entity\_quantity} attribute specifies how many entities there are, which are then logically grouped within a container.

\noindent(2) \textbf{Container} represents the grouping or possession of entities as indicated in the MWP, similar to the definition in~\citep{opedal-etal-2023-world}. 
For example, in Figure~\ref{fig:fig2}, Faye is a container that possesses 88 colorful flowers.
A container is identified by attributes \texttt{container\_name}, \texttt{container\_type}, \texttt{attr\_name} and \texttt{attr\_type}. 
The \texttt{container\_name} describes the container's name as stated in the MWP, while \texttt{container\_type} defines its category for visualization. 
In Figure~\ref{fig:fig2}, ``Faye'' is the \texttt{container\_name} and ``girl'' is the \texttt{container\_type}. The \texttt{attr\_name} and \texttt{attr\_type} are optional attributes that provide additional contextual details of the container. 

\noindent(3) \textbf{Operation} represents mathematical or logical relationships between containers. 
In addition to basic arithmetic operations such as addition, subtraction, multiplication, and division, we incorporate additional operations including surplus, comparison and unit transformation.
These operations enable us to cover 94.4\% of Grade 1-3 MWPs in the ASDiv dataset~\citep{asdiv2020}.
Operations are denoted as:
\vspace{-5pt}
\begin{eqnarray}
    \texttt{operation}(\texttt{container1},\, \texttt{container2},\, \hspace{1cm}\\\nonumber
    \texttt{result\_container})
\end{eqnarray}

The {\bf final \ac{VL}} is a composition of the solution expression $E_{solution}$ and the operations. Thus, container1 and container2 in eq. 1 can themselves be operations, enabling nested operations and supporting hierarchical representations for more complex MWPs. 
We use identical attributes for container1, container2, and result\_container to ensure consistency and ease LLM interpretation. 
For example, in Figure~\ref{fig:fig2}, an inner subtraction operation is performed between container Faye and Mike, and the resulting value is divided by container bouquet through an outer division operation.
We show the comparison of our \ac{VL} with other semantic parsing methods of MWPs in Table~\ref{tab:tab12}.
Our method has the most comprehensive arithmetic coverage (+, -, *, /, surplus, >, <) and is among the only two approaches that can handle multiple-order MWPs.

\subsection{Visual Design}\label{sec:design_element}

In this section, we describe how elements from the \ac{VL} are visualized. Our design, informed by an exploratory study with five primary school math teachers (Section~\ref{sec:design_expert}), is inspired by the bar model~\cite{hoven2007singapore} and Noyon’s modular design~\citep{saquib2021constructing} (Section~\ref{sec:literature}).

\paragraph{Container with Entity:} 
Inspired by Noyon’s modular design and the bar model’s structure, we depict containers as rectangles enclosing visualized entities. For quantities over ten, a single entity is shown with its number overlaid, consistent with Twinkl datasets~\citep{twinkl}.
The attributes \texttt{container\_name} and \texttt{container\_type} are visualized as a small icon accompanied by text above the container rectangle, as shown in Figure~\ref{fig:fig2}. 
Additionally, if \texttt{attr\_name} and \texttt{attr\_type} have non-empty values, they are displayed as the icon alongside the container icon.
\paragraph{Operation:} \label{sec:operation}
As informed by exploratory study (see Section~\ref{sec:design_expert}), we visualize operations using two visual variations: ``Formal'' and ``Intuitive''. The ``Formal'' variation represents operations using mathematical symbols (e.g. ``+'', ``-'', ``×'', ``÷'') accompanied by text, as shown in Figure~\ref{fig:fig2}. More examples are in Appendix~\ref{sec:example_formal}.

In the ``Intuitive'' variation, each operation is represented through a specific visual arrangement, we present high level description below, with more details in Appendix~\ref{sec:detail_operation}.

\noindent{\bf $\bullet$ Addition:} Containers in the addition operation are enclosed in a large rectangle (see Figure~\ref{fig:fig14}).

\noindent{\bf $\bullet$ Subtraction}: The minuend container is visualized first, with the subtracted entities crossed out (see Figure~\ref{fig:fig15}). 

\noindent{\bf $\bullet$ Multiplication}: The multiplicand container is repeated to represent multiplication (see Figure~\ref{fig:fig16}). For special area computing problems, it is depicted as a single entity with dimensions matching the MWP's width and length (see Figure~\ref{fig:fig23}).

\noindent{\bf $\bullet$ Division}: The division operation is visualized as the post-division state, with multiple entity rectangles representing groups enclosed within a larger rectangle (see example in Figure~\ref{fig:fig17} and \ref{fig:fig18}).

\noindent{\bf $\bullet$ Surplus}: Similar to division, but the surplus entity is visualized separately (e.g., see Figure~\ref{fig:fig0}). 

\noindent{\bf $\bullet$ Comparison}: This operation involves comparing different entities by visualizing them on a balance scale. Each entity is placed on one side of the scale (see example in Figure~\ref{fig:fig20}).

\noindent{\bf $\bullet$ Unit Transformation}:  The unit transformation operation is for questions that involve changes in measurement units. We adopt a purple bubble above each entity to display its value in the transformed unit (see example in Figure~\ref{fig:fig21}).

Finally, for MWPs with multiple operations, we follow these visualization rules for each operation and dynamically combine them to form the overall expression tree (see Figure~\ref{fig:fig22}).

\subsection{From Visual Language to Visual}\label{sec:render}

We convert our Visual Languages (VLs) into visuals using dedicated rendering programs. Each entity in \ac{VL} is mapped to a visual icon from an SVG dataset, while preserving the operations and relationships between the containers. To achieve this, we convert the \ac{VL} into a tree structure that captures the hierarchical relationships between operations and containers. 
We traverse the tree to compute the relative positions of each container in the visual based on its attributes (such as \texttt{entity\_quantity}) and the layout corresponding to the involved operations (see Section~\ref{sec:operation}). 
The overall process produces a global layout plan for rendering. 
Finally, we traverse the tree, assigning a corresponding SVG icon for each ``type'' attribute (\texttt{entity\_type}, \texttt{container\_type}, and \texttt{attr\_type}) and render the complete visual based on the global layout plan. 
Note that the attributes in \texttt{result\_container} are only used in ``Intuitive'' visual generation. 
The complete algorithm is presented in Algorithm~\ref{alg:vl_to_svg}.

\subsection{Validating Designs with Teachers}\label{sec:design_expert}
{\bf Co-Designing Visuals with Teachers:} We conducted an exploratory study with five experienced primary school math teachers (Grades 1–3; demographics in Table~\ref{tab:tab7}) who regularly use visuals to teach MWPs. 
During the study, participants evaluated six alternative visual designs for the same set of MWPs. These alternatives were inspired by bar models~\cite{hoven2007singapore} and Noyon’s modular design framework~\citep{saquib2021constructing}, and were developed to explore variation in how entities and mathematical relationships are visually represented. Further details on the design rationale, study protocol, and results are provided in Appendix~\ref{sec:detail_explore_study}.

\paragraph{Participants Recognize Our Design's Value for Teaching} \label{sec:post-task}
Our exploratory study results (Tables~\ref{tab:tab1} and \ref{tab:tab2}) confirmed that teachers perceive our visuals as effective in clearly conveying the central ideas of MWP, reducing unnecessary cognitive load, and enhancing student engagement.
We asked participants to rate our visual design on a 7-point Likert scale (7 being the highest). Every participant awarded a perfect 7.0 for both ``usefulness for teaching'' and ``likelihood of frequent use in class,'' and the average score for ``helpfulness for student understanding of MWPs'' was 6.8. These ratings indicate that our design is pedagogically meaningful.
\paragraph{Suggestions for Refining the Visual Representation}
Participants highlighted two key insights: the ``Formal'' design, which incorporates math symbols, best enhances the clarity of mathematical expressions, while the ``Intuitive'' design best improves student engagement and reduces unnecessary cognitive load. Their feedback on how quantities should be represented led to refinements in our approach. Consequently, our final design offers two variations: ``Formal'' design emphasizing clarity and ``Intuitive'' design tailored for engagement and optimization of cognitive load. 

\subsection{Evaluation Criteria for Generated Visuals} \label{sec:eval_criteria}
 
After discussions with five math teachers, we established the following criteria to evaluate our generation approach in reproducing our design.

\noindent\textbf{(i) Accuracy} measures how accurately the quantity of entities and relationships between entities in the visual reflect the MWP. This criterion is crucial in education as it is important for students to learn accurate information~\citep{metzger2003college, goldin2001systems}.

\noindent\textbf{(ii) Completeness} evaluates whether all elements necessary for solving the MWP --- including entities, quantities, mathematical relationships, and contextual cues that affect problem interpretation --- are present in the visual.
This criterion is vital in education, as teachers should provide complete and necessary information to learners~\citep{crosby2000amee}.

\noindent\textbf{(iii) Clarity} measures how easily students can interpret the visual without confusion or ambiguity. This includes clear distinctions between entities, appropriate use of labels and unambiguous spatial arrangements.
Clarity is important in math teaching, as it supports effective learning~\citep{metzger2003college, goldin2001systems}.

\noindent\textbf{(iv) Cognitive Load Optimization} assesses whether the visual minimizes unnecessary cognitive load caused by distractions or redundant details that do not contribute to problem-solving. Minimizing unnecessary cognitive load is crucial since learners' working memory can process only a few elements at a time~\cite{paul2002}.

\section{Visual Dataset Generation}
In this section, we describe the process of generating a visual dataset from MWPs.

\subsection{MWP Data Source} 
We select the ASDiv dataset~\citep{asdiv2020} as our source of MWPs as it covers a diverse range of problem types and includes Grade-level annotations for each question. 
We collect 1,268 MWPs suitable for our \sys{} framework, constituting 94.4\% of the Grade 1–3 MWPs in ASDiv.

\subsection{Dataset Creation}\label{sec:dataset_creation}
In this section, we explain our dataset creation process. First, we manually wrote 30 \ac{VL} examples that serve as in-context demonstrations for LLMs.
Using these examples, we prompt the o1-mini model~\citep{gpt4o1-mini} to generate the remaining \ac{VL} for our collected MWPs. 
The prompt is shown in Appendix~\ref{sec:prompt_vl}. 
For each generated \ac{VL}, we automatically retrieve the entities for visualization and manually collect the corresponding SVG icons of these entities from multiple sources~\citep{svgrepoRepoFree, iconfont2025, svgen2025, kaggleIcons, huggingfaceUmuthopeyildirimsvgen500kDatasets, pexels}. These SVG icons are then combined with the \ac{VL} to render a total of 1,903 visuals --- comprising 1,268 ``Formal'' visuals and 635 ``Intuitive'' visuals. 
Finally, two researchers manually validate each rendered visual and its associated \ac{VL} to ensure it accurately represents the corresponding MWP. The process, including SVG collection and manual verification, required approximately 160 hours of dedicated effort. Table~\ref{tab:tab10} provides an overview of our annotated dataset and comparisons with other math pedagogical visual datasets.

\section{Results and Analysis}
In this section, we aim to address the following experimental questions regarding \sys{}:

\noindent\circled{1} How does the choice of generation framework affect the quality of the generated visuals?

\noindent\circled{2} How does incorporating the solution expression of an MWP impact the generation results?

\noindent\circled{3} How does fine-tuning on synthesized visual dataset enhance model performance in generating pedagogical meaningful visuals?


\subsection{Experiment Design}\label{sec:exp_design}
To assess various strategies for producing visuals, we conduct two sets of experiments: one to evaluate how effectively LLMs generate \ac{VL}, and another to compare our \sys{} framework with the latest TTI models for generating visuals.
 
\textbf{Evaluating LLMs for Generating Visual Language}
We create a test set of 257 \ac{VL} instances using stratified sampling based on ``Grade'' (e.g., Grade 1) and ``Question Type'' (e.g., addition) from our annotated dataset.
We compare two recent LLMs with strong reasoning capabilities: OpenAI o3-mini~\citep{o3-mini} and Gemini 2.0 Flash~\citep{googleGeminiFlash}, to see how accurately they can generate \ac{VL}. We provide both models with prompts in Appendix~\ref{sec:prompt_vl} and vary whether we include solution expressions in the prompt to test the effect on generation quality. We measure performance by computing: (1) Logic Match Ratio: 
The proportion of generated VLs whose underlying logical structure --- specifically, the set of operations (e.g., addition, subtraction) and associated entity quantities --- exactly matches the annotated ground truth VLs.
(2) Edit Distance: The average distance between generated and ground-truth \ac{VL}. We compute this using Zhang–Shasha tree edit distance algorithm~\citep{zhang1989simple}, implemented through the zss package\footnote{https://github.com/timtadh/zhang-shasha}.
Each VL is parsed into a hierarchical tree structure, where nodes represent either operations or containers. 
Operation nodes serve as parent nodes and have their input containers or nested operations as child nodes. 
The tree edit distance is then computed as the minimum number of node insertions, deletions, or substitutions required to transform the generated VL tree into the ground-truth VL tree.

\textbf{Evaluating Methods for Generating Visuals}
For evaluating different methods of generating visuals, we conduct a two-stage assessment.
In the first stage, we perform initial human evaluation using two test sets (Formal and Intuitive), each containing 24 visuals. These visuals are stratifiedly sampled based on "Grade" and "Question Type" from our annotated dataset.

We evaluate two state-of-the-art TTI models, DALLE-3~\citep{dalle3} and Recraft-V3~\citep{recraftv3}, using prompts detailed in Appendix~\ref{sec:prompt_visual} while experimenting with both prompts with and without solution expressions.
We compare the visuals generated by DALLE-3 and Recraft-V3 with those rendered from the
\ac{VL} generated by o3-mini and Gemini 2.0 Flash, alongside ground-truth visuals from our annotated dataset. Two researchers independently evaluated each visual based on the criteria described in Section~\ref{sec:eval_criteria}.

To validate results of initial evaluation, we selected the best-performing TTI model and \sys{} with the best LLM for an expanded human evaluation using the same settings. For this phase, we created two additional test sets (Formal and Intuitive), each with 72 visuals.

\subsection{Results of Visual Language Evaluation} \label{sec:eval_llm}

In the upper part of Table~\ref{tab:tab4}, we show evaluation results as the \ac{VL} is generated by various methods. 
The o3-mini with solution expression input (i.e. when mathematical expression for solving the MWP is given) achieves a logic match ratio of 96.89\%, indicating close alignment with the ground truth in operations and entity quantities. 
The Gemini-2-flash model with solution expression records the lowest edit distance, suggesting its generated \ac{VL} closely matches the ground truth attribute values.

Within the same model, including the solution expression reduces the edit distance while increasing the logic match ratio of the generated \ac{VL}. However, advanced models like o3-mini can still achieve a 91\% logic match without expression.

We present additional results in Appendix~\ref{sec:compare_vl} comparing the solving accuracy of the o3-mini model with existing graph-based MWP solvers~\citep{bin-etal-2023-non,hu-etal-2023-generation}. The results show that o3-mini achieves the highest accuracy across the evaluated datasets.
\begin{table}[h]
    \centering
    \small
    \begin{tabular}{p{3cm}p{1.5cm}p{1.7cm}}
     \toprule
      \textbf{Criterion}   & \textbf{Edit Dist$\downarrow$}  & \textbf{LM Ratio$\uparrow$} \\
      \midrule
     o3-mini(E) & 2.82 & \textbf{96.89}\\
     o3-mini & 2.90 & 91.05\\
      gemini-2-flash(E) & \textbf{2.67} & 90.27\\
      gemini-2-flash & 2.96 & 72.76\\
    \midrule
    ft\_llama-3.1-8B(E) & \textbf{2.28} & \textbf{89.50} \\
    ft\_llama-3.1-8B & 2.52 & 80.54 \\
    zs\_llama-3.1-8B(E) & 4.67 & 1.95 \\
    zs\_llama-3.1-8B & 4.47 & 3.11 \\
    \hline
    \end{tabular}
    \caption{Visual Language Generation Results: E indicates generation with the solution expression. Scores are averaged over 257 \(VL\) instances per method.}
    \label{tab:tab4}
    \vspace{-15pt}
\end{table}
\begin{table*}
\centering
\scriptsize
\begin{tabular}{p{.1cm}p{2cm}M{1cm}M{1cm}M{1cm}M{1cm}M{1cm}M{1cm}M{1cm}M{1cm}|M{1cm}}
\toprule
& \textbf{Method} & \multicolumn{2}{c}{\textbf{Accuracy}} & \multicolumn{2}{c}{\textbf{Completeness}} & \multicolumn{2}{c}{\textbf{Clarity}} & \multicolumn{2}{c}{\textbf{Cog Load Opt}} & \textbf{Avg.} \\
\midrule
& & \textbf{Formal} & \textbf{Intuitive} & \textbf{Formal} & \textbf{Intuitive} & \textbf{Formal} & \textbf{Intuitive} & \textbf{Formal} & \textbf{Intuitive} & \\
\midrule
\multirow{8}{*}{\rotatebox[origin=c]{90}{prompting}} 
& o3-mini(E) & \textbf{4.92} & 4.58 & \textbf{5.00} & \textbf{4.67} & 4.88 & 4.67 & \textbf{4.96} & 4.54 & \textbf{4.78} \\
& o3-mini & 4.83 & 4.54 & 4.96 & 4.50 & \textbf{5.00} & 4.67 & \textbf{4.96} & 4.42 & 4.74 \\
& gemini-2-flash(E) & 4.79 & 4.50 & 4.92 & 4.57 & 4.96 & \textbf{4.79} & \textbf{4.96} & \textbf{4.65} & 4.77 \\
& gemini-2-flash & 4.54 & \textbf{4.62} & 4.58 & 4.57 & 4.79 & 4.67 & 4.75 & 4.61 & 4.64 \\
& recraft-v3(E) & 3.33 & 2.96 & 3.75 & 3.62 & 3.75 & 4.00 & 3.63 & 3.96 & 3.63 \\
& recraft-v3 & 3.26 & 2.96 & 3.5 & 3.33 & 3.54 & 3.75 & 3.54 & 3.92 & 3.48 \\
& dalle-3(E) & 2.96 & 3.04 & 3.21 & 3.42 & 2.54 & 2.33 & 2.54 & 2.50 & 2.82 \\
& dalle-3 & 2.79 & 2.96 & 2.83 & 3.33 & 2.12 & 2.29 & 2.17 & 2.46 & 2.62 \\
\midrule
\multirow{8}{*}{\rotatebox[origin=c]{90}{fine-tuning}} 
& ft\_llama-3.1-8B(E) & \textbf{4.79} & \textbf{4.83} & \textbf{4.83} & \textbf{4.83} & \textbf{4.83} & \textbf{4.83} & \textbf{4.83} & \textbf{4.83} & \textbf{4.83} \\
& ft\_llama-3.1-8B & 4.58 & \textbf{4.83} & 4.63 & \textbf{4.83} & 4.67 & \textbf{4.83} & 4.67 & \textbf{4.83} & 4.73 \\
& zs\_llama-3.1-8B(E) & 1.25 & 1.33 & 1.25 & 1.33 & 1.33 & 1.33 & 1.29 & 1.33 & 1.31 \\
& zs\_llama-3.1-8B & 1.08 & 1.00 & 1.04 & 1.00 & 1.17 & 1.00 & 1.17 & 1.00 & 1.06 \\
& ft\_flux.1-dev(E) & 3.21 & 2.62 & 3.38 & 3.38 & 3.38 & 3.12 & 3.50 & 3.33 & 3.24 \\
& ft\_flux.1-dev & 3.12 & 2.21 & 3.33 & 3.38 & 3.33 & 3.17 & 3.29 & 3.25 & 3.14 \\
& zs\_flux.1-dev(E) & 3.13 & 2.50 & 3.21 & 2.83 & 3.33 & 3.25 & 3.42 & 3.63 & 3.16 \\
& zs\_flux.1-dev & 3.13 & 2.42 & 3.21 & 2.83 & 3.33 & 3.25 & 3.42 & 3.63 & 3.15 \\
\midrule\midrule
\multirow{4}{*}{\rotatebox[origin=c]{90}{exp. eval}} 
& o3-mini(E) & \textbf{4.97} & \textbf{4.96} & \textbf{5.00} & \textbf{4.97} & 4.94 & \textbf{4.94} & \textbf{4.96} & 4.96 & \textbf{4.96} \\
& recraft-v3(E) & 2.65 & 3.00 & 3.57 & 3.82 & 3.58 & 3.76 & 3.18 & 3.29 & 3.36 \\
& ft\_llama-3.1-8B(E) & 4.93 & 4.92 & 4.92 & 4.92 & \textbf{4.99} & 4.92 & \textbf{4.96} & \textbf{4.97} & 4.94 \\
& ft\_flux.1-dev(E) & 2.49 & 2.53 & 2.60 & 2.64 & 3.67 & 3.54 & 3.89 & 3.92 & 3.16 \\
\bottomrule
\end{tabular}
\caption{Human Evaluation of Visual Representations: In the upper and middle parts of the table, 48 visuals (24 Formal, 24 Intuitive) were evaluated with scores averaged from two researchers on a 1–5 scale. In the lower part (expanded evaluation), 144 visuals (72 Formal, 72 Intuitive) were further evaluated with the best performing models. (E) indicates use of the solution expression as input; ``ft'' denotes a fine-tuned model and ``zs'' a zero-shot model.}
\label{tab:tab5}
\vspace{-5pt}
\end{table*}
\subsection{Results of Visual Evaluation} \label{sec:eval_method}
The upper part of Table~\ref{tab:tab5} shows evaluation results for visuals generated by different methods. Based on these, we selected o3-mini(E) and recraft-v3(E) for the expanded human evaluation, with results in the lower part of Table~\ref{tab:tab5}. These results confirm the trends observed in our initial human evaluation. Our key findings are as follows:

\textbf{\sys{} Scores Highly on All Criteria}
The \sys{} framework, equipped with the latest LLMs, outperforms other TTI models across all criteria, demonstrating its capability to generate accurate visuals aligned with our design. The o3-mini model performs best on the Formal dataset, while the Gemini-2-flash model achieves better results on the Intuitive dataset. The scores for the Formal dataset are consistently higher across criteria compared to the Intuitive dataset. This discrepancy may be due to the Intuitive dataset containing slightly more complex questions for converting to \ac{VL}. However, the score difference remains relatively small, around 0.4.

\textbf{Solution Expression Increases Performance}
Within the same model, including the solution expression as input increases performance in most cases, possibly because it offers a structured representation of the MWP that helps the model understand the mathematical relationships between containers.

\subsection{Fine-tuning for Visual Generation}\label{sec:fine-tune}
In this section, we evaluate the effectiveness of fine-tuning LLMs and TTI models using annotated dataset. We fine-tuned the LLMs using 80\% of the annotated data, while for the TTI models, we fine-tuned Formal models with 80\% of the Formal data and Intuitive models with 80\% of the Intuitive data (details in Appendix~\ref{sec:fine-tune_detail}). Specifically, we fine-tuned two LLMs, Llama-3.1-8B~\citep{dubey2024llama} and Mistral-7B-v0.3~\citep{mistral7bv03}, as well as two TTI models, Flux.1-dev~\citep{FLUX1dev} and Stable Diffusion-3.5-large~\citep{esser2024scaling}. For each model, we fine-tuned two versions: one using dataset with solution expression input and one without. Results for Llama-3.1-8B are presented in Tables~\ref{tab:tab4} and \ref{tab:tab5}, for Flux.1-dev in Table~\ref{tab:tab5}, and for the other models in Tables~\ref{tab:tab8} and \ref{tab:tab9}. 

As shown in the lower part of Table~\ref{tab:tab4}, the Llama model fine-tuned with expression achieves the lowest edit distance among all models, significantly reducing the edit distance compared to its zero-shot version. It also achieves a logic match ratio comparable to the latest LLMs and higher than that of the model fine-tuned without expression input.

The middle section of Table~\ref{tab:tab5} shows that the visuals generated by \sys{} with the Llama model fine-tuned with the expression achieve scores comparable to those of the latest LLMs across all criteria. Similarly, the Flux model fine-tuned with the expression performs comparably to the latest TTI models. In every instance, models fine-tuned on datasets with expression outperform those without expression. Our expanded human evaluation (see lower section of Table~\ref{tab:tab5}) further validates these findings.

\begin{table*}
\centering
\scriptsize
\begin{tabular}{p{2.1cm}M{0.9cm}M{0.9cm}M{0.9cm}M{0.9cm}M{0.9cm}M{0.9cm}M{0.9cm}M{0.9cm}M{0.9cm}M{0.9cm}}
\toprule
\textbf{Method} & \multicolumn{2}{c}{\textbf{Quantity Err}} & \multicolumn{2}{c}{\textbf{Relation Err}} & \multicolumn{2}{c}{\textbf{Struct Misalign}} & \multicolumn{2}{c}{\textbf{ Miss Visual Item}} & \multicolumn{2}{c}{\textbf{Miss Contex Cue}}\\
\midrule
 & \textbf{Formal} & \textbf{Intuitive} & \textbf{Formal} & \textbf{Intuitive} & \textbf{Formal} & \textbf{Intuitive} & \textbf{Formal} & \textbf{Intuitive} & \textbf{Formal} & \textbf{Intuitive}\\
\midrule
ft\_flux.1-dev(E) & 0.72 & 0.74 & 0.85 & \textbf{0.81} & \textbf{0.35} & \textbf{0.23} & \textbf{0.44} & 0.30 & 0.57 & \textbf{0.49} \\
zs\_flux.1-dev(E) & 0.77 & 0.78 & 0.92 & 0.85 & 1.00 & 1.00 & 0.74 & 0.66 & 0.62 & 0.60 \\
recraft-v3(E)    & \textbf{0.41} & \textbf{0.38} & \textbf{0.82} & \textbf{0.81} & 0.64 & 0.94 & \textbf{0.44} & \textbf{0.18} & \textbf{0.35} & 0.50 \\
\bottomrule
\end{tabular}
\caption{Statistical Results for Qualitative Analysis: For each method, 192 visuals (96 Formal and 96 Intuitive) were evaluated, with each score representing the ratio of corresponding error.}
\label{tab:tab11}
\vspace{-10pt}
\end{table*}
\subsection{Qualitative Analysis on TTI Models and Discussion}
To identify and understand common errors in visuals generated by TTI models, we performed a qualitative analysis.
We use thematic analysis to identify recurring error patterns. This process involved two phases: an initial exploration with 120 visuals to identify error types and then a systematic evaluation of visuals using these categories with 576 visuals generated by three representative methods.
The error types include: (1) \textbf{Quantity Error}: an incorrect number of entities; (2) \textbf{Relation Error}: incorrect mathematical relationships between containers; (3) \textbf{Structural Misalignment}: visuals that do not align structurally with our design, featuring misaligned elements or disorganized groupings; (4) \textbf{Missing Visual Item}: visuals missing necessary entities for solving MWP; and (5) \textbf{Missing Contextual Cue}: visuals lacking essential contextual cues for solving MWP.
We present examples representing each error type in Appendix~\ref{sec:error_type}.
Table~\ref{tab:tab11} summarizes the ratio of each error type, with key findings discussed below. Detailed breakdowns by Grade and operation type are provided in Appendix~\ref{sec:stat_results}.

\textbf{Fine-tuning Improves Structural Alignment and Entities Inclusion}
Table~\ref{tab:tab11} shows that fine-tuning the Flux model significantly reduces structural misalignment errors compared to the zero-shot model. The fine-tuned model generated visuals align to our design by consistently representing containers as rectangles encompassing entities. In contrast, while the zero-shot version generally represents quantities accurately as numbers, it often fails to properly visualize the corresponding entities. 
We also observe that fine-tuning significantly reduces missing visual item errors compared to the zero-shot model. In the zero-shot setting, models often generate only numerical representations and fail to include visual items necessary for solving the corresponding MWPs. For example, Figure~\ref{fig:fig29} was generated by the zero-shot Flux.1-dev model with expression input; the figure contains only the equation and omits essential items such as ``penny'' and ``nickel''. In contrast, the fine-tuned Flux.1-dev with expression input more reliably visualizes these items, as shown in Figure~\ref{fig:fig31}, resulting in visuals that are both more interpretable and engaging. 
According to Table~\ref{tab:tab13}, fine-tuning is especially effective in reducing structural misalignment and missing visual item errors in higher-Grade problems (Grades 2 and 3), where underlying solution expression and language become more complex.
Overall, fine-tuning decreases error rates across all evaluated categories.

\textbf{Relation Errors Remain a Severe Problem}
Despite improvements from fine-tuning, all models continue to exhibit high relation error rates --- ranging from 0.82 to 0.92 in Formal and 0.81 to 0.85 in Intuitive --- indicating that visualizing mathematical relationships remains a persistent challenge. 
These errors typically arise when models apply incorrect operations or depict relationships in ambiguous ways.
For example, Figure~\ref{fig:fig27}, generated by the Recraft-v3 model with expression input, incorrectly represents a multiplication problem as an addition scenario and depicts an incorrect number of bees. This misrepresentation not only introduces numerical inaccuracies but also distorts the intended reasoning structure of the problem --- potentially confusing learners about which operation to apply.
These findings suggest that current TTI models struggle to visually represent the relationships between quantities --- particularly in problems involving comparison and surplus operations, where understanding depends on how groups relate to one another rather than on individual values. A breakdown of error rates by operation type is provided in Table~\ref{tab:tab14}.
While existing work has explored methods for generating precise numerical quantities in visuals~\citep{binyamin2024make}, further research is needed to develop techniques that effectively visualize mathematical relationships.

\section{Conclusion}
This work introduces \sys{}, an automatic framework for generating scalable and pedagogically meaningful visuals from MWP text descriptions. \sys{} leverages a tree-based visual language and a structured visual design space --- developed in collaboration with math teachers --- to effectively capture the essential mathematical relationships within MWPs. Using \sys{}, we generated and annotated a dataset of 1,903 visuals and evaluated state-of-the-art Text-to-Image (TTI) models on their ability to produce visuals that align with our design. We further demonstrated that fine-tuning these models on our dataset improves the quality of visual generation. While our results represent a promising step toward the automated generation of pedagogically meaningful visuals, challenges remain in directly generating such visuals with current TTI models. Future work will explore more scalable and flexible generation frameworks and further refine our visual design to better support educational outcomes.

\section*{Limitations}
\textbf{(i) Scope of Representation}
\sys{} is currently limited to math word problems involving the seven operations defined in this paper (addition, subtraction, multiplication, division, surplus, comparison, unit transformation). Although our framework can handle MWPs that require multiple operations, the solution must be representable in a single expression. 
While \sys{} does not currently support math word problems involving multiple interdependent equations, its modular design makes such extensions feasible. In principle, such problems could be decomposed into a sequence of intermediate VLs, each of which can be visualized individually within the existing framework. Extending \sys{} to support such problems---particularly those involving variable substitution, elimination, or symbolic manipulation---represents a promising direction for future research.


\noindent\textbf{(ii) Language Restriction}
Our study focuses solely on MWPs written in English. While \sys{} should, in principle, be applicable to similar problems in other languages, adapting the system for multilingual support remains an avenue for future exploration.

\noindent\textbf{(iii) Predefined Visual Style and Input Requirements}
Despite achieving 94.4\% coverage of Grade 1-3 MWPs in the ASDiv dataset~\citep{asdiv2020}, \sys{} relies on a predefined visual style and requires a SVG dataset of entity icons as input. Although this controlled approach ensures the pedagogical validity of visuals and is an effective strategy given current model capabilities, it inherently limits generation flexibility. Future research may explore more versatile frameworks, such as adapting advanced Text-to-Image models, to generate pedagogically valuable visuals without predefined styles.
\section{Acknowledgements}
This project was made possible by ETH AI Center Doctoral Fellowships to Junling Wang, with partial support from the ETH Zurich Foundation. We thank Yifan Hou for the insightful discussion about the design of the TTI model evaluation experiments. We are also grateful to Prof. Dennis Komm for helping us advertise and recruit participants for the human evaluation study. Additionally, the authors wish to thank the reviewers, members of the LRE Lab and PEACH Lab at ETH Zurich, and the participants in the human evaluation experiments.

\bibliographystyle{acl_natbib}

\appendix
\label{sec:appendix}
\section{Visual Language Details}
\subsection{Example of Visual Language}\label{sec:example_vl}
In the MWP description ``Jake picked up three apples in the morning...'' the container1 could be specified as \texttt{entity\_name: apple}, \texttt{entity\_type: apple}, \texttt{entity\_quantity: 3}, \texttt{container\_name: Jake}, \texttt{container\_type: boy}, \texttt{attr\_name: morning}, \texttt{attr\_type: morning}. These additional attributes are not fixed and may vary according to different interpretations.
\subsection{Comparison of Visual Language with Other MWP Works} \label{sec:compare_vl}
We show the comparison of our Visual Language with other semantic parsing methods of MWPs in Table~\ref{tab:tab12}.

To evaluate the MWP-solving accuracy of the o3-mini model, we conducted additional experiments comparing it with two graph-based MWP solvers~\citep{bin-etal-2023-non,hu-etal-2023-generation} on the ASDiv\citep{asdiv2020} and MAWPS\citep{koncel-kedziorski-etal-2016-mawps} datasets. For o3-mini, we employed a prompting-based approach to directly generate the solution. In contrast, the graph-based methods were trained using 80\% of each dataset, with the remaining 20\% used as a test set---shared across all models for consistency. We adopted the default configurations provided in the official codebases of the graph-based solvers. The accuracy results, summarized in Table~\ref{tab:tab15}, show that o3-mini outperforms the other methods and achieves the highest accuracy on both datasets.

\begin{table*}
\scriptsize
\centering
\begin{tabular}{p{3cm}|M{2cm}|M{3cm}|M{2cm}|M{2cm}}
\hline
\textbf{Work} & \textbf{Arithmetic Coverage} & \textbf{Conceptual Coverage} & \textbf{Semantic Granularity}  & \textbf{Problem Depth} \\
\hline
Visual Language (ours) & (+, -, ×, ÷, surplus, >,<) & Transfer, Rate, Comparison, Part-whole, Surplus, Unit Transformation, Multiple Steps & Concepts \& equations & multiple-order MWP\\
\hline
\citep{opedal-etal-2023-world} & (+, -, ×, ÷) & Transfer, Rate, Comparison, Part-whole & World model & first-order MWP \\
\hline
\citep{hosseini2014learning} & (+, -) & Transfer & World model & first-order MWP \\
\hline
\citep{mitra2016learning} & (+, -) & Transfer, Comparison (add), Part-whole & Concepts \& equations & first-order MWP \\
\hline
\citep{roy2018} & (+, -, ×, ÷) & Transfer, Rate, Comparison, Part-whole, Concepts \& equations & Concepts \& equations & multiple-order MWP \\
\hline
\end{tabular}

\caption{Comparison of our Visual Language approach with existing semantic parsing methods for MWPs.}
\label{tab:tab12}
\end{table*}
\begin{table}[h]
    \centering
    \scriptsize
    \begin{tabular}{p{3cm}p{2.3cm}p{1cm}}
     \toprule
      \textbf{Method}   & \textbf{ASDiv(Grade 1-3)} & \textbf{MAWPS} \\
      \midrule
      		
      Non-autoregressive MWP~\citep{bin-etal-2023-non}  & 0.67	& 0.91 \\
      Generation-based Deductive~\citep{hu-etal-2023-generation} & 0.79 &	0.92 \\
      o3-mini	& \textbf{0.97} & \textbf{0.97} \\

      \bottomrule       
    \end{tabular}
    \caption{Comparison of o3-mini with two graph-based MWP solvers. Values indicate accuracy on different datasets.}
    \label{tab:tab15}
\end{table}

\section{Example of Visuals}
\subsection{Example of Formal Visual}\label{sec:example_formal}
\begin{figure}[h!]
    \centering
    \includegraphics[width=0.5\textwidth]{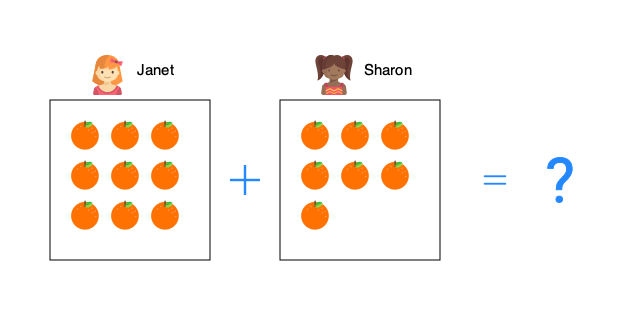}
    \caption{Example of addition operation in Formal design (Intuitive version: Figure~\ref{fig:fig14}). Corresponding MWP: Janet has nine oranges, and Sharon has seven oranges. How many oranges do Janet and Sharon have together?
    }
    \label{fig:fig6}
\end{figure}

\begin{figure}[h!]
    \centering
    \includegraphics[width=0.5\textwidth]{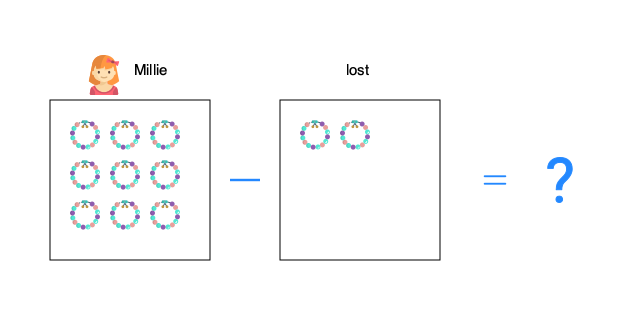}
    \caption{Example of subtraction operation in Formal design (Intuitive version: Figure~\ref{fig:fig15}). Corresponding MWP: Millie had 9 bracelets. She lost 2 of them. How many bracelets does Millie have left?
    }
    \label{fig:fig7}
\end{figure}

\begin{figure}[h!]
    \centering
    \includegraphics[width=0.5\textwidth]{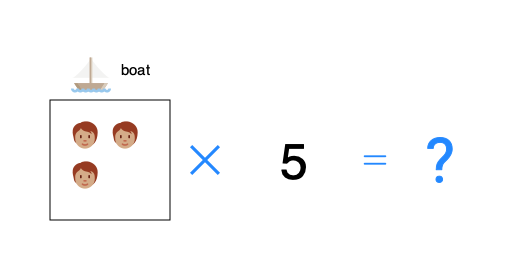}
    \caption{Example of multiplication operation in Formal design (Intuitive version: Figure~\ref{fig:fig16}). Corresponding MWP: 5 boats are in the lake. Each boat has 3 people. How many people are on boats in the lake?
    }
    \label{fig:fig8}
\end{figure}

\begin{figure}[h!]
    \centering
    \includegraphics[width=0.5\textwidth]{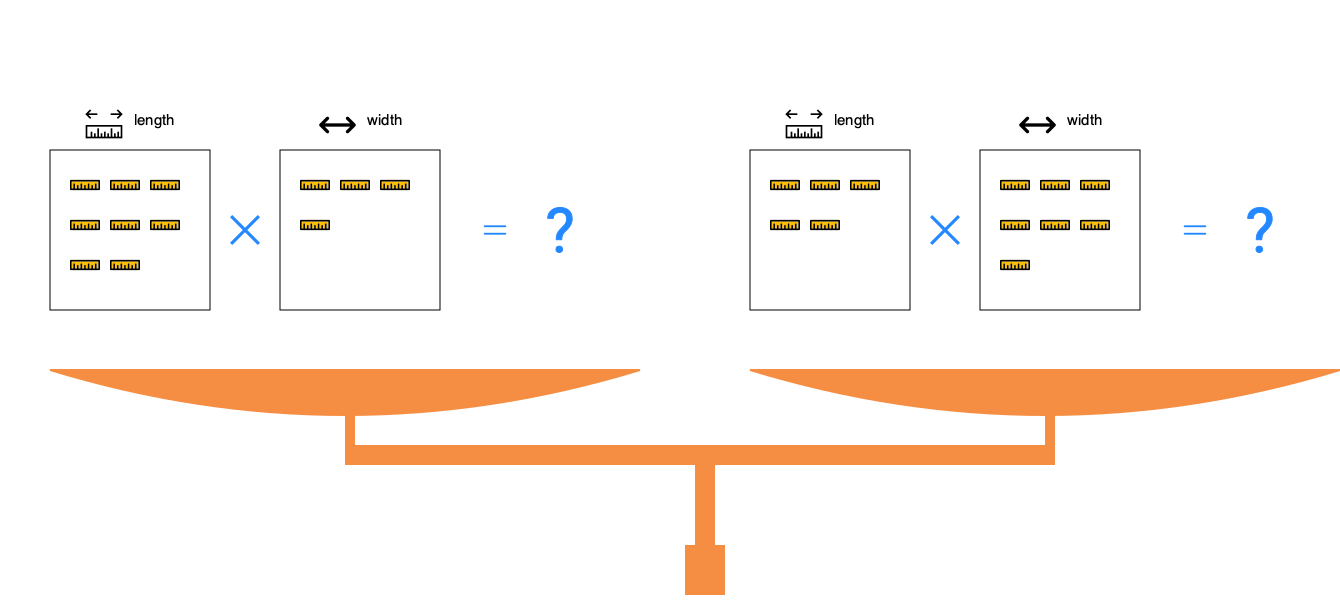}
    \caption{Example of area operation (a special type of multiplication operation) in Formal design (Intuitive version: Figure~\ref{fig:fig23}). We use the ruler icon to represent measurement units like feet, meters, etc. Corresponding MWP: Rug A is 8 feet by 4 feet, and Rug B is 5 feet by 7 feet. Which rug should Mrs. Hilt buy if she wants the rug with the biggest area?
    }
    \label{fig:fig25}
\end{figure}

\begin{figure}[h!]
    \centering
    \includegraphics[width=0.5\textwidth]{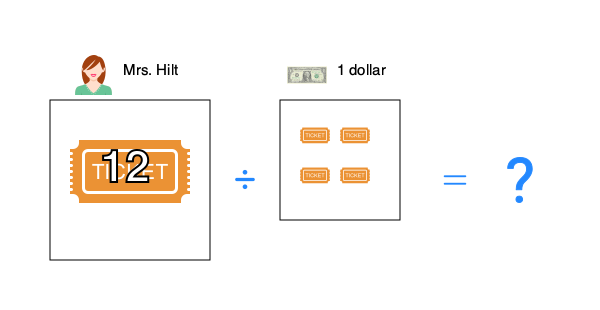}
    \caption{Example of division operation in Formal design (Intuitive version: Figure~\ref{fig:fig18}). Corresponding MWP: Mrs. Hilt bought carnival tickets. The tickets cost \$1 for 4 tickets. If Mrs. Hilt bought 12 tickets, how much did she pay?
    }
    \label{fig:fig9}
\end{figure}

\begin{figure}[h!]
    \centering
    \includegraphics[width=0.5\textwidth]{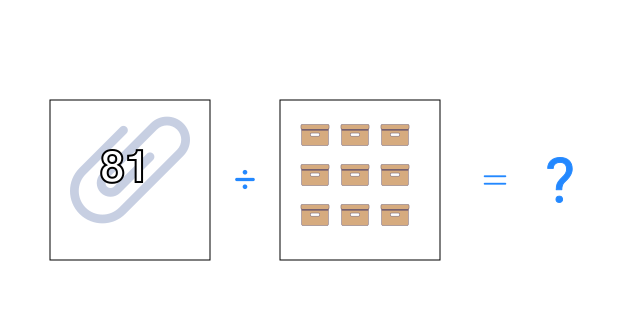}
    \caption{Example of division operation in Formal design (Intuitive version: Figure~\ref{fig:fig17}). It represents visuals of a division operation in an MWP, asking for the quantity per group. Corresponding MWP: Lexie's younger brother helped pick up all the paper clips in Lexie's room. He was able to collect 81 paper clips. If he wants to distribute the paper clips in 9 boxes, how many paper clips will each box contain?
    }
    \label{fig:fig24}
\end{figure}

\begin{figure}[h!]
    \centering
    \includegraphics[width=0.5\textwidth]{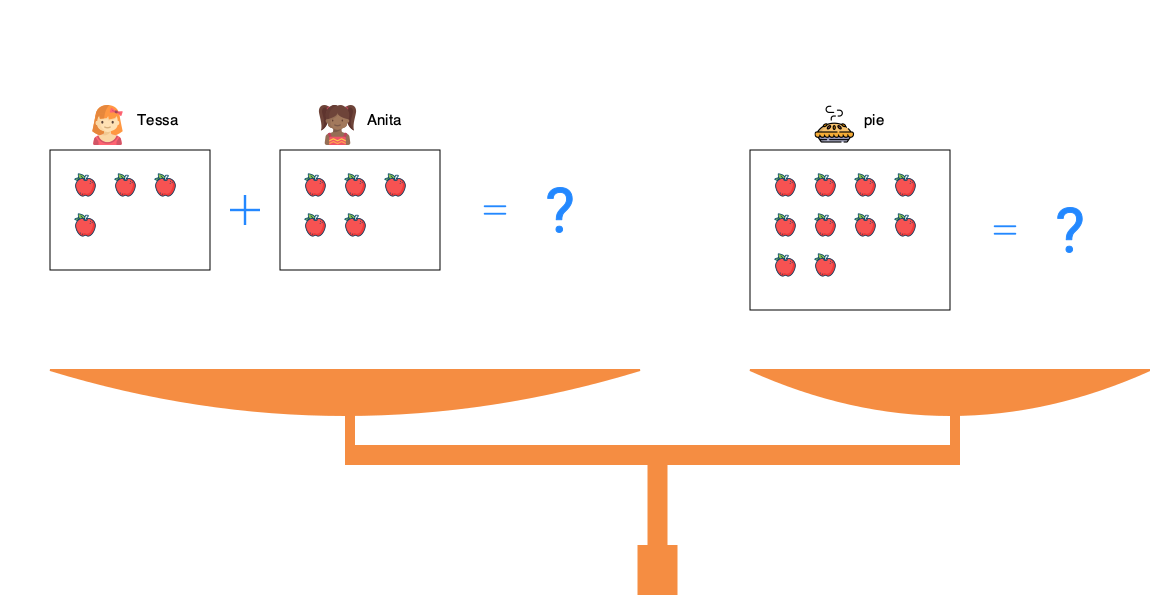}
    \caption{Example of comparison operation in Formal design (Intuitive version: Figure~\ref{fig:fig20}). Corresponding MWP: Tessa has 4 apples. Anita gave her 5 more. She needs 10 apples to make a pie. Does she have enough to make a pie?
    }
    \label{fig:fig11}
\end{figure}

\begin{figure}[h!]
    \centering
    \includegraphics[width=0.5\textwidth]{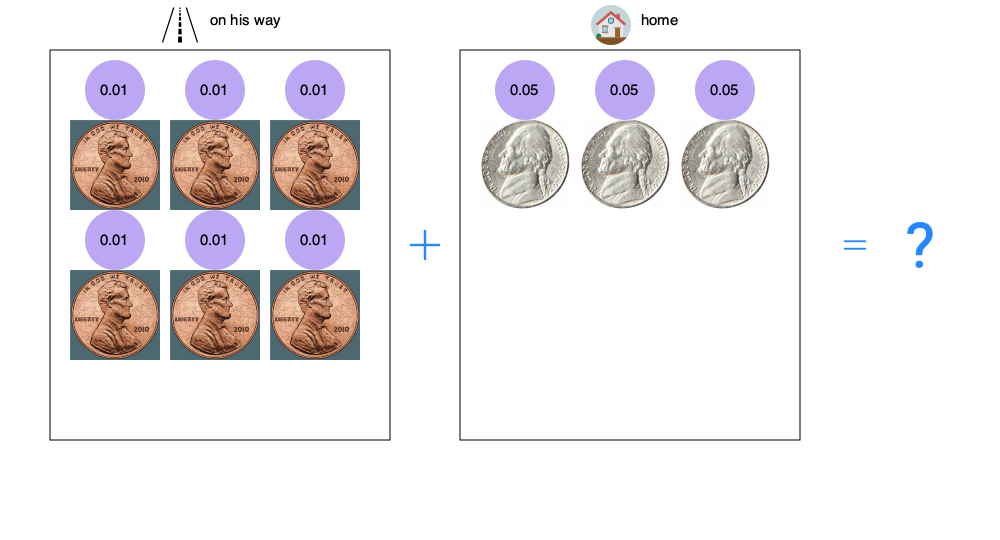}
    \caption{Example of unit transformation operation in Formal design (Intuitive version: Figure~\ref{fig:fig21}). Corresponding MWP: Charles found 6 pennies on his way to school. He also had 3 nickels already at home. How much money does he now have in all?
    }
    \label{fig:fig12}
\end{figure}

\begin{figure}[h!]
    \centering
    \includegraphics[width=0.5\textwidth]{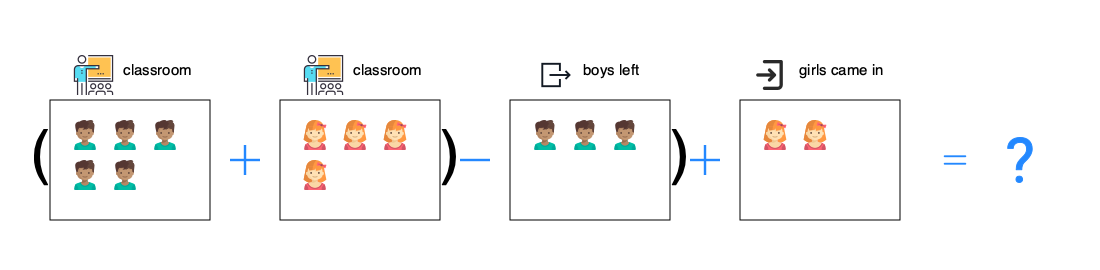}
    \caption{Example of multiple steps operation in Formal design (Intuitive version: Figure~\ref{fig:fig22}). Corresponding MWP: There are 5 boys and 4 girls in a classroom. After 3 boys left the classroom, another 2 girls came in the classroom. How many children were there in the classroom in the end?
    }
    \label{fig:fig13}
\end{figure}

We provide examples of ``Formal'' visuals in Figures~\ref{fig:fig6} to \ref{fig:fig13}.

\subsection{Example of Intuitive Visual}\label{sec:example_intuitive}
\begin{figure}[h!]
    \centering
    \includegraphics[width=0.5\textwidth]{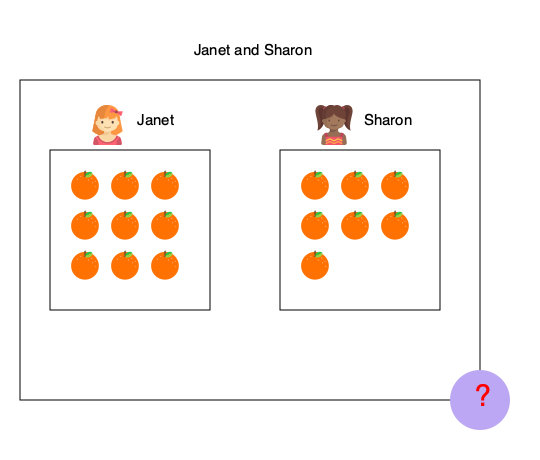}
    \caption{Example of addition operation in Intuitive design (Formal version: Figure~\ref{fig:fig6}). Corresponding MWP: Janet has nine oranges and Sharon has seven oranges. How many oranges do Janet and Sharon have together?
    }
    \label{fig:fig14}
\end{figure}

\begin{figure}[h!]
    \centering
    \includegraphics[width=0.3\textwidth]{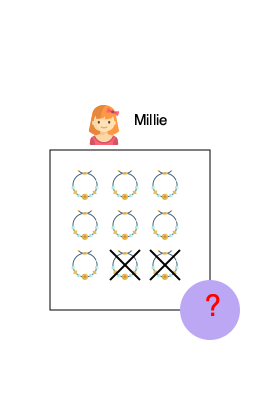}
    \caption{Example of subtraction operation in Intuitive design (Formal version: Figure~\ref{fig:fig7}). Corresponding MWP: Millie had 9 bracelets. She lost 2 of them. How many bracelets does Millie have left?
    }
    \label{fig:fig15}
\end{figure}

\begin{figure}[h!]
    \centering
    \includegraphics[width=0.5\textwidth]{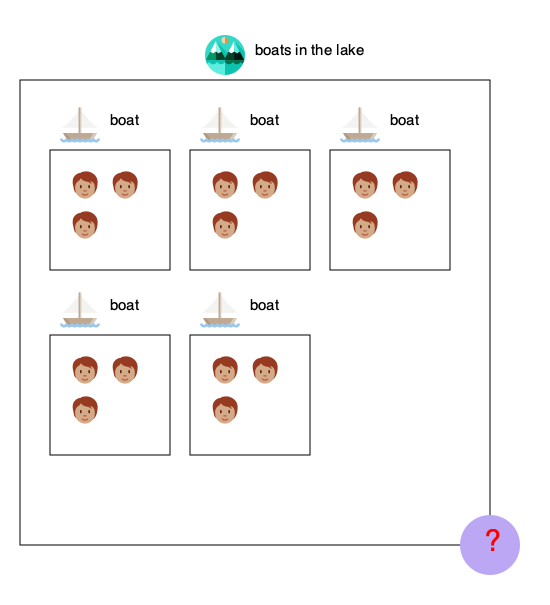}
    \caption{Example of multiplication operation in Intuitive design (Formal version: Figure~\ref{fig:fig8}). Corresponding MWP: 5 boats are in the lake. Each boat has 3 people. How many people are on boats in the lake?
    }
    \label{fig:fig16}
\end{figure}

\begin{figure}[h!]
    \centering
    \includegraphics[width=0.5\textwidth]{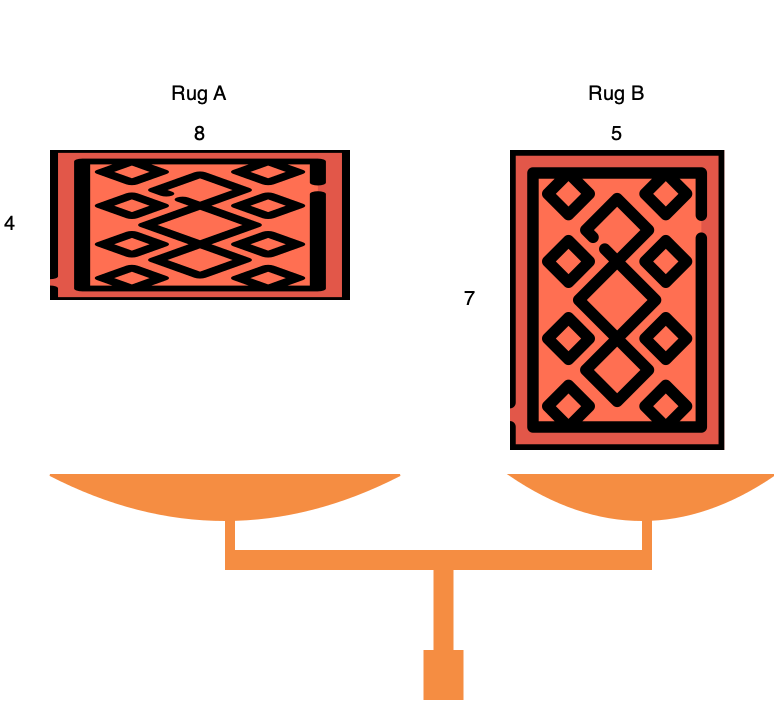}
    \caption{Example of area operation (a special type of multiplication operation) in Intuitive design (Formal version: Figure~\ref{fig:fig25}). Corresponding MWP: Rug A is 8 feet by 4 feet, and Rug B is 5 feet by 7 feet. Which rug should Mrs. Hilt buy if she wants the rug with the biggest area?
    }
    \label{fig:fig23}
\end{figure}

\begin{figure}[h!]
    \centering
    \includegraphics[width=0.5\textwidth]{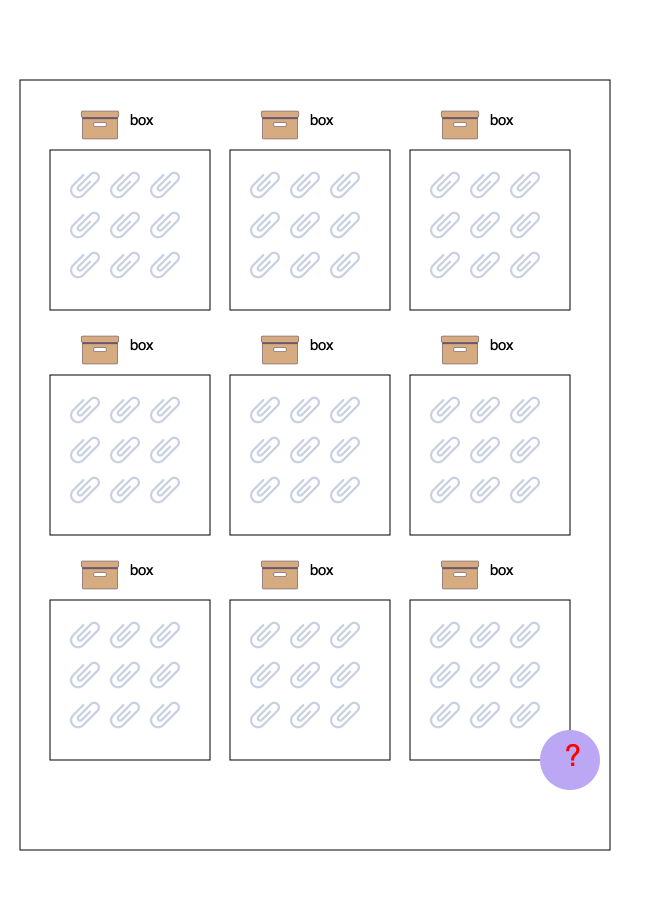}
    \caption{Example of division operation in Intuitive design (Formal version: Figure~\ref{fig:fig24}). It represents visuals of a division operation in an MWP, asking for the quantity per group. Corresponding MWP: Lexie's younger brother helped pick up all the paper clips in Lexie's room. He was able to collect 81 paper clips. If he wants to distribute the paper clips in 9 boxes, how many paper clips will each box contain?
    }
    \label{fig:fig17}
\end{figure}

\begin{figure}[h!]
    \centering
    \includegraphics[width=0.5\textwidth]{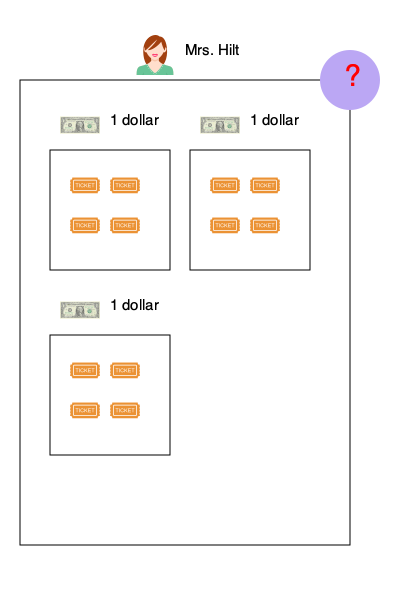}
    \caption{Example of division operation in Intuitive design (Formal version: Figure~\ref{fig:fig9}). It represents visuals of a division operation in an MWP, asking for the number of groups. Corresponding MWP: Mrs. Hilt bought carnival tickets. The tickets cost \$1 for 4 tickets. If Mrs. Hilt bought 12 tickets, how much did she pay?
    }
    \label{fig:fig18}
\end{figure}

\begin{figure}[h!]
    \centering
    \includegraphics[width=0.5\textwidth]{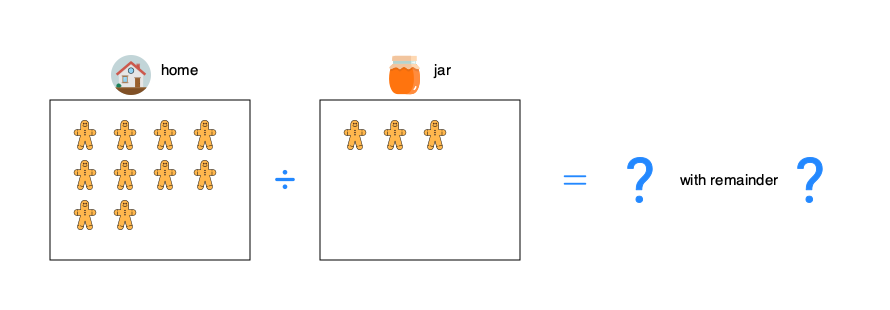}
    \caption{Example of surplus operation in Intuitive design (Intuitive version: Figure~\ref{fig:fig0}). Corresponding MWP: At home, Marian made 10 gingerbread cookies which she will distribute equally in tiny glass jars. If each jar is to contain 3 cookies each, how many cookies will not be placed in a jar?
    }
    \label{fig:fig19}
\end{figure}

\begin{figure}[h!]
    \centering
    \includegraphics[width=0.5\textwidth]{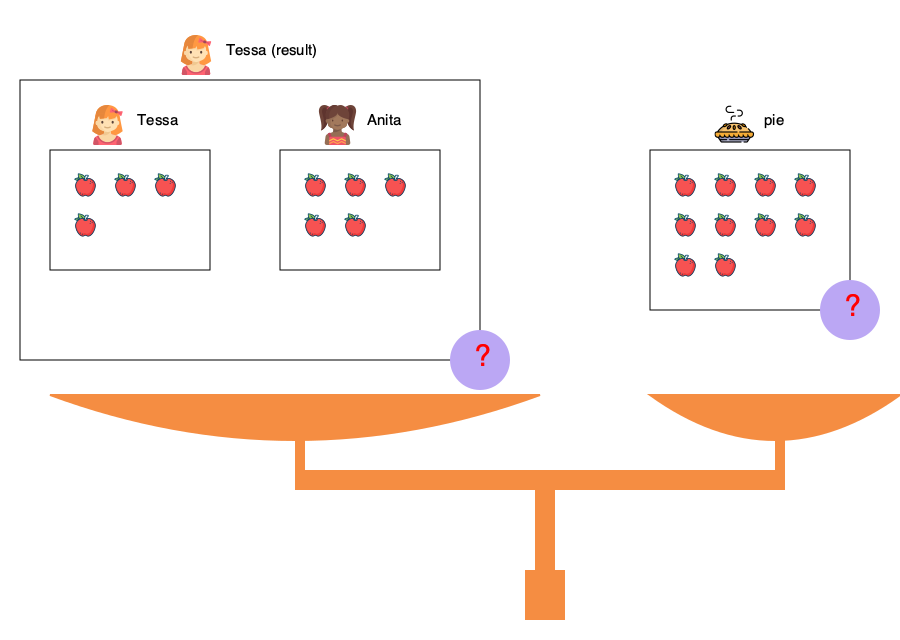}
    \caption{Example of comparison operation in Intuitive design (Formal version: Figure~\ref{fig:fig11}). Corresponding MWP: Tessa has 4 apples. Anita gave her 5 more. She needs 10 apples to make a pie. Does she have enough to make a pie?
    }
    \label{fig:fig20}
\end{figure}

\begin{figure}[h!]
    \centering
    \includegraphics[width=0.5\textwidth]{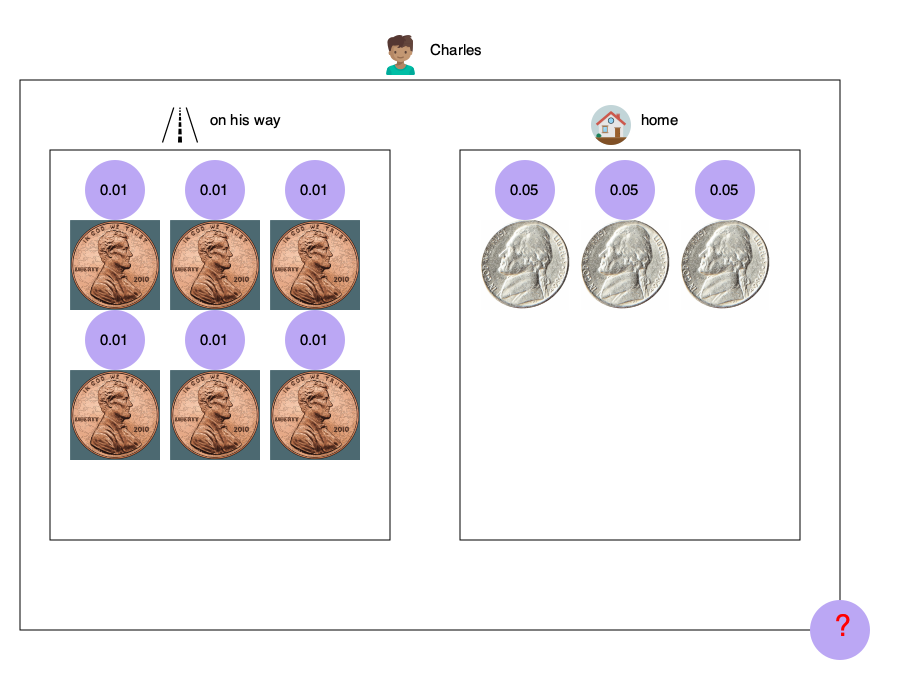}
    \caption{Example of unit transformation operation in Intuitive design (Formal version: Figure~\ref{fig:fig12}). Corresponding MWP: Charles found 6 pennies on his way to school. He also had 3 nickels already at home. How much money does he now have in all?
    }
    \label{fig:fig21}
\end{figure}

\begin{figure}[h!]
    \centering
    \includegraphics[width=0.5\textwidth]{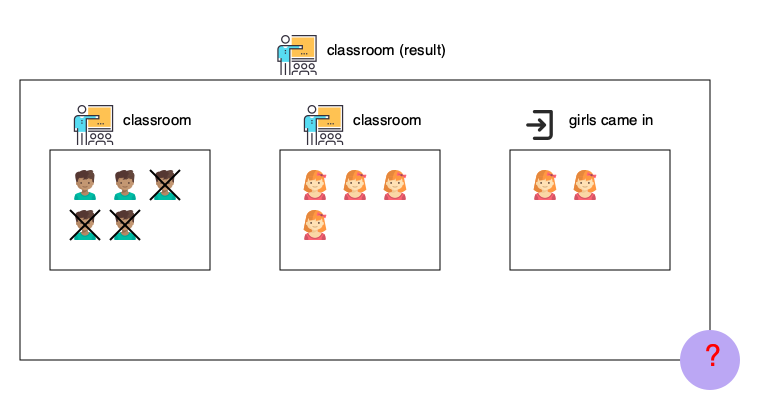}
    \caption{Example of multiple steps operation in Intuitive design (Formal version: Figure~\ref{fig:fig13}). Corresponding MWP: There are 5 boys and 4 girls in a classroom. After 3 boys left the classroom, another 2 girls came in the classroom. How many children were there in the classroom in the end?
    }
    \label{fig:fig22}
\end{figure}

We provide examples of ``Intuitive'' visuals in Figures~\ref{fig:fig14} to \ref{fig:fig22}.


\subsection{Representative Visual Examples for Each Error Type} \label{sec:error_type}
\begin{figure}[h!]
    \centering
    \includegraphics[width=0.5\textwidth]{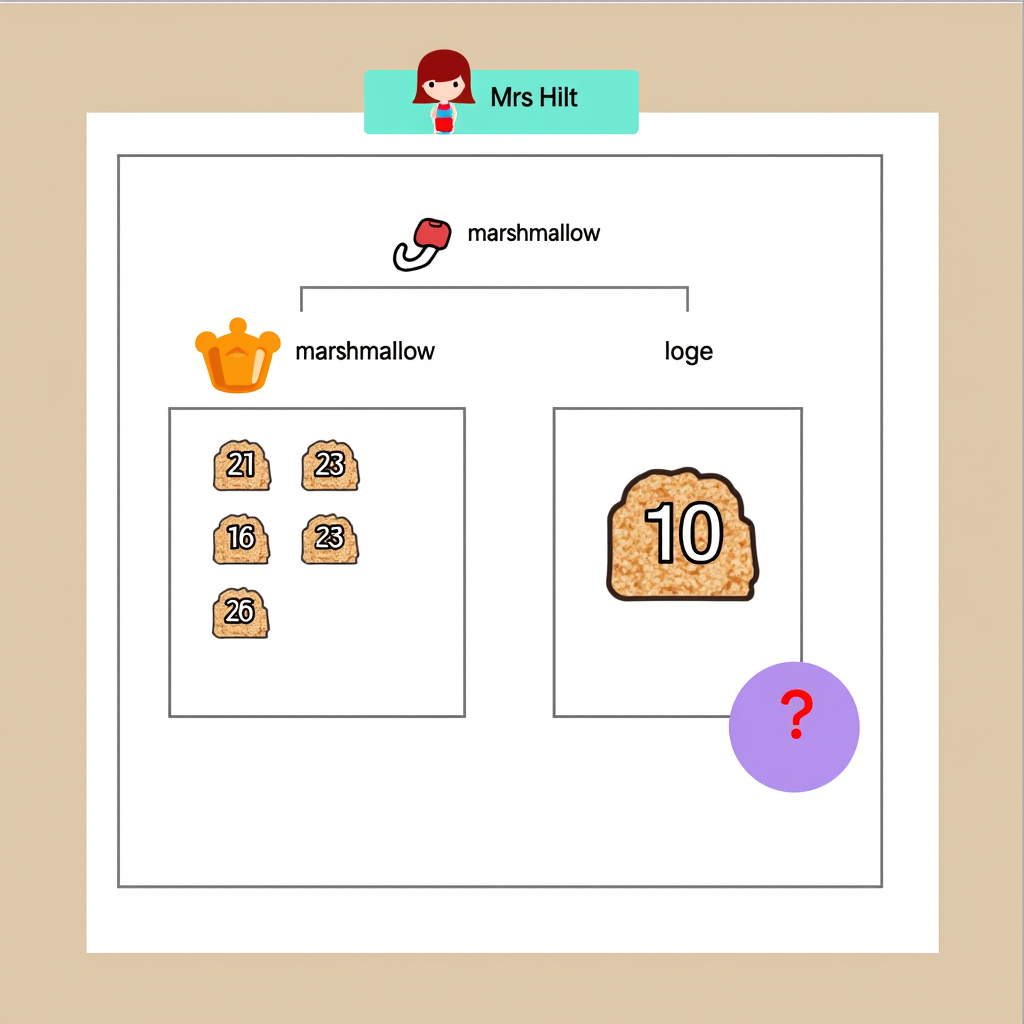}
    \caption{Example visual with quantity error. Corresponding MWP: Mrs. Hilt made 5 Rice Krispie Treats. She used 8 large marshmallows and 10 mini marshmallows. How many marshmallows did she use altogether?
    }
    \label{fig:fig26}
\end{figure}

\begin{figure}[h!]
    \centering
    \includegraphics[width=0.5\textwidth]{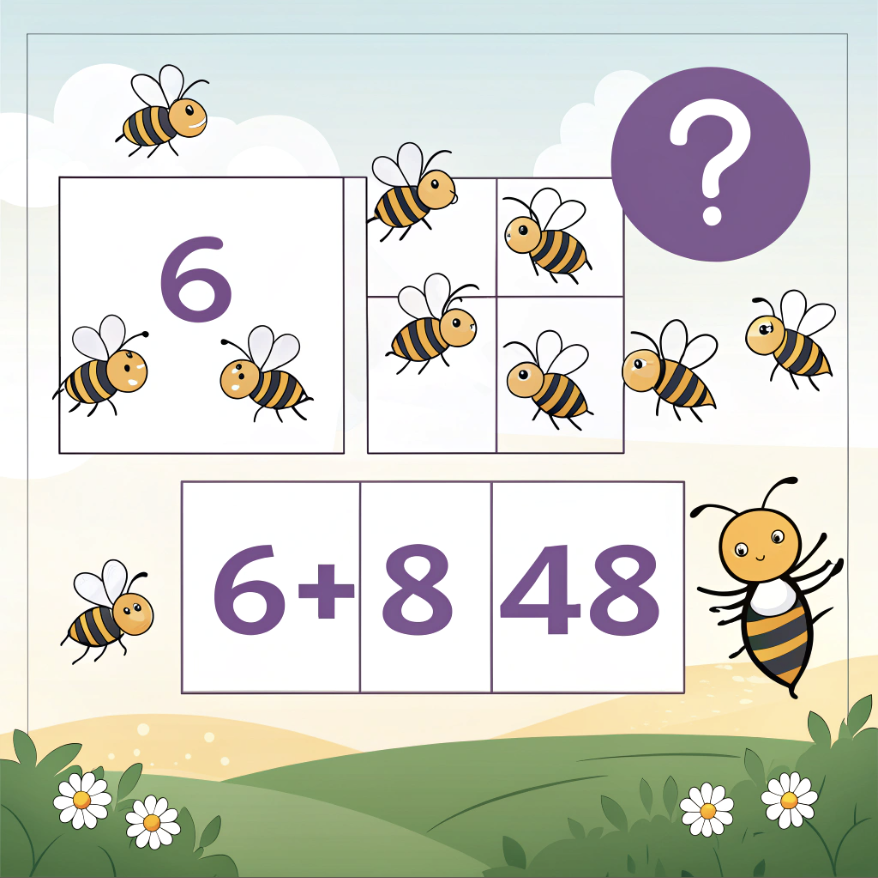}
    \caption{Example visual with relation error. Corresponding MWP: A bee has 6 legs. How many legs do 8 bees have?
    }
    \label{fig:fig27}
\end{figure}

\begin{figure}[h!]
    \centering
    \includegraphics[width=0.5\textwidth]{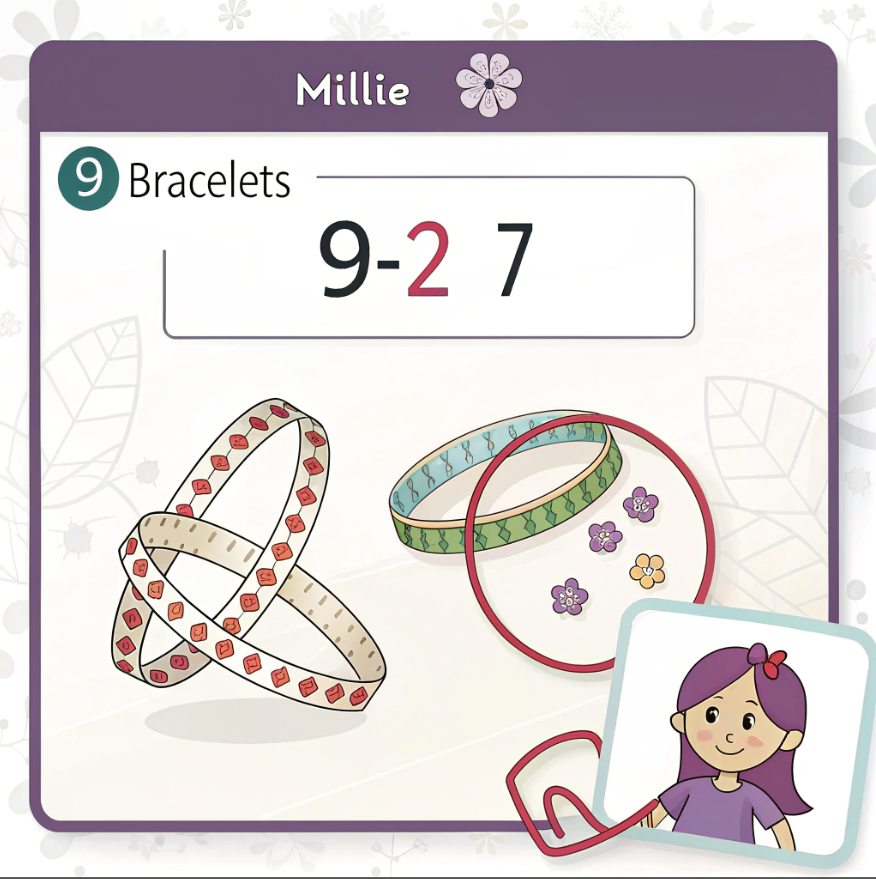}
    \caption{Example visual with structure misalignment error. Corresponding MWP: Millie had 9 bracelets. She lost 2 of them. How many bracelets does Millie have left?
    }
    \label{fig:fig28}
\end{figure}

\begin{figure}[h!]
    \centering
    \includegraphics[width=0.5\textwidth]{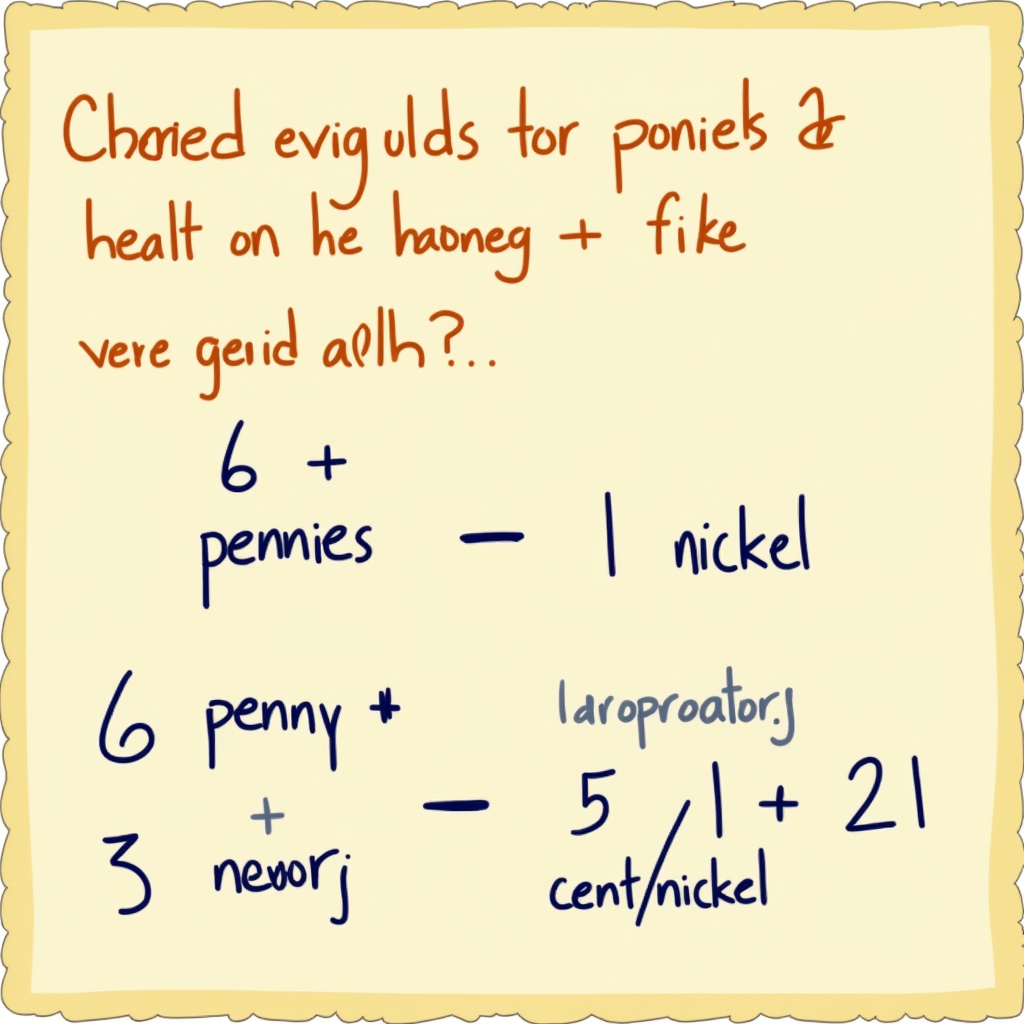}
    \caption{Example visual with missing visual item error. Corresponding MWP: Charles found 6 pennies on his way to school. He also had 3 nickels already at home. How much money does he now have in all? This example is generated by zero-shot Flux.1-dev with solution expression, the corresponding example generated by fine-tuned Flux.1-dev with solution expression is shown in Figure~\ref{fig:fig31}.
    }
    \label{fig:fig29}
\end{figure}

\begin{figure}[h!]
    \centering
    \includegraphics[width=0.5\textwidth]{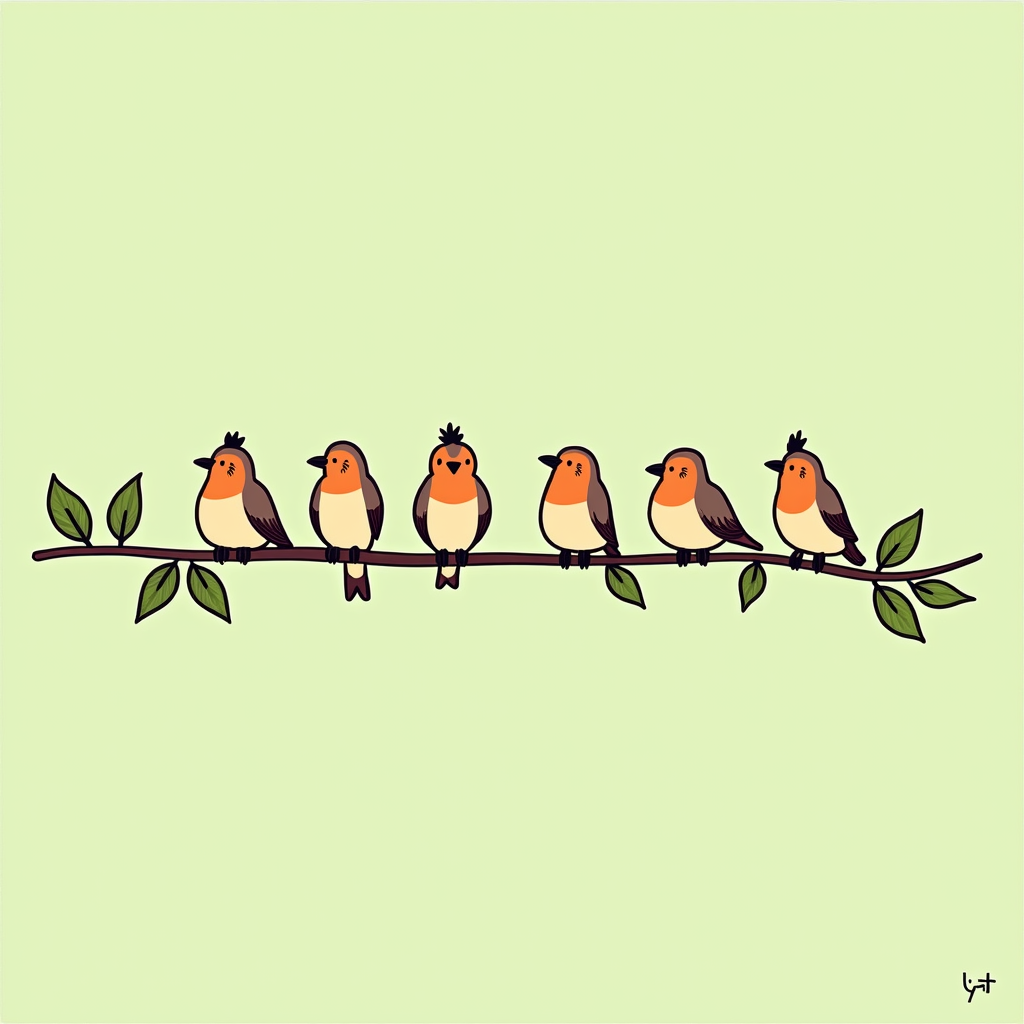}
    \caption{Example visual with missing contextual cues error. Corresponding MWP: 4 birds are sitting on a branch. 1 flies away. How many birds are left on the branch?
    }
    \label{fig:fig30}
\end{figure}

We provide examples visuals of each error category in Figures~\ref{fig:fig26} to \ref{fig:fig30}. 
\begin{figure}[h!]
    \centering
    \includegraphics[width=0.5\textwidth]{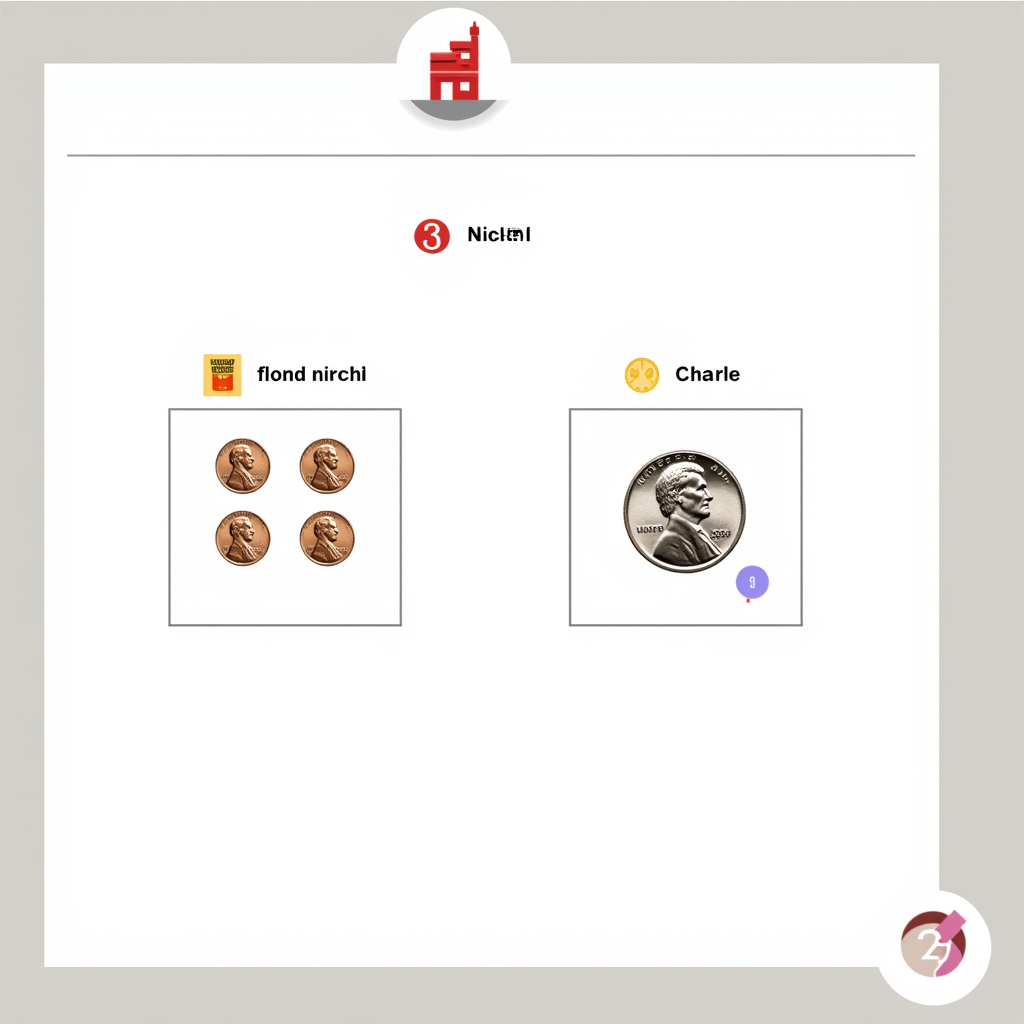}
    \caption{Corresponding Example Generated by Fine-tuned Flux.1-dev with solution expression. Corresponding MWP: Charles found 6 pennies on his way to school. He also had 3 nickels already at home. How much money does he now have in all?
    }
    \label{fig:fig31}
\end{figure}

\section{Details of Exploration Study}\label{sec:detail_explore_study}

\subsection{Participants’ Demographics}
We recruited primary school math teachers through Prolific~\citep{prolificProlificEasily} and paid them 15 USD per hour, which is adequate given the participants' country of residence.
We present the participants' demographics in Table~\ref{tab:tab7}.
\begin{table}[h]
    \centering
    \begin{tabular}{cp{2cm}p{1cm}p{2cm}}
     \toprule
      \textbf{PID}   & \textbf{Language of Teaching} & \textbf{Age} & \textbf{Gender}\\
      \midrule
      
       1 & English  & 52 & Male\\
       2 & English  & 45 & Male\\
       3 & English  & 35 & Female\\
       4 & English  & 44 & Female\\
       5 & English  & 37 & Female\\

      \bottomrule       
    \end{tabular}
    \caption{Participants’ Demographics: We recruited five primary school math teachers who teach Grades 1–3 through Prolific. All teachers consider themselves experienced educators in using visuals to teach MWPs.}
    \label{tab:tab7}
\end{table}

\subsection{Study Protocol}\label{sec:study_protocol}
Our study obtained ethical approval and collected consent forms from each participant. During the study, participants were first introduced to the background of the study. They then completed four sessions, as described below. The entire study ranged from 1.5h to 2h.

In the first session, participants were asked to indicate their preference between two visual approaches: (1) using multiple visuals, where each visual represents one sentence of the MWP, or (2) using a single visual to represent the entire MWP.

In the second session, we presented six design variations to the participants. These variations differed based on two design choices:

\begin{enumerate}
\itemsep0em 
    \item How Quantities Are Visualized:
    \begin{itemize}
        \item \textbf{Abstract}: Quantities are represented as text from the MWP.
        \item \textbf{Hybrid}: A single item is visualized with a label at the bottom-right corner indicating its quantity.
        \item \textbf{Visual}: Items are directly drawn in quantities matching their number.
    \end{itemize}
    \item How Operations Are Visualized:
    \begin{itemize}
        \item \textbf{Formal}: Mathematical operations are represented using standard symbols (e.g., +, -, ×, ÷).
        \item \textbf{Intuitive}: Operations are visualized using specific arrangements for each operation, as described in Section~\ref{sec:operation}.
    \end{itemize}
\end{enumerate}

By combining the three approaches for quantities and the two approaches for operations, we created six unique design variations. Each variation was introduced to the participants and their feedback was sought based on the following criteria: (1) \textbf{Clarity}: The extent to which the visual design clearly represents the math word problem. (2) \textbf{Engagement}: Whether the visual design helps improve student engagement. (3) \textbf{Cognitive Load}: Whether the visual design avoids introducing unnecessary cognitive load for students.

We asked participants to complete a questionnaire after reviewing each design and collected their suggestions for improving the respective design variations. We randomized the presentation order of the design variations to minimize order effects.

In the third session, we aimed to gather feedback on our ``Intuitive'' design, which visualizes different operations. The design details are presented in Section~\ref{sec:operation}. We used the same criteria as in session two and asked participants to complete a questionnaire after reviewing each operation design, collecting their suggestions for improvement. We also randomized the presentation sequence of designs in session three.

In session four, we discussed with each participant the criteria to use for analyzing the subsequently generated visuals. This evaluation focused not on the design itself but on how effectively our generation approach could reproduce the intended design. After completing all the sessions, we asked participants to complete a post-task questionnaire assessing the pedagogical value of our visual design. The results are presented in Section~\ref{sec:post-task}.
\subsection{Additional Results} \label{sec:additional_results}
\paragraph{Single Visual is Preferred for Clarity and Simplicity} Most of the participants (4) prefered the single visual design than multiple visual per MWP. They mentioned single visual have better clarity and is explicit enough for simple MWP for Grade 1-3 students. 
\paragraph{Paticipants' Suggestions on Design Decisions}
The results of study session two are presented in Table~\ref{tab:tab1}. Participants noted that the use of math symbols in the 'Formal' design enhances clarity, while the 'Visual' and 'Intuitive' designs increase engagement and reduce unnecessary cognitive load. However, they also mentioned that the purple circle with a quantity inside caused confusion for learners. They recommended displaying the quantity directly on the visual item and reserving the circle exclusively for question marks. Based on participants' feedback, we refined the designs and developed the final version, which includes two variations: the 'Formal' design using math symbols and the 'Intuitive' design featuring specific arrangements for different operations. More details about our final design are provided in Section~\ref{sec:design_element}.

\begin{table}[h]
    \centering
    \small
    \begin{tabular}{cccc}
     \toprule
      \textbf{Design}   & \textbf{Clarity} & \textbf{Engagement} & \textbf{Cog Load Opt}\\
      \midrule
      
       AF & \textbf{5.0}  & 5.4 & 4.6\\
       AI &  3.6  & 4.6 & 3.0  \\
       HF & 3.0  & 3.6 & 1.6\\
       HI & 2.0  & 4.2 & 1.4\\
       VF & 4.6  & 5.6 & 3.4\\
       VI & 4.8 & \textbf{6.0} & \textbf{5.2}\\
    \midrule
    \end{tabular}
    \caption{Results of exploratory study session 2. In ``Design'' column, ``A'' represents ``Abstract''; ``H'' means ``Hybrid''; ``V'' means ``Visual''; ``F'' means ``Formal''; ``I'' means Intuitive. Different combinations reflect different designs, which we discuss in Appendix~\ref{sec:additional_results}. All scores are on a 7-point Likert scale, where higher values indicate better performance. Four participants indicated that, after slight modifications to the question mark, the clarity score of the AI design would be 7.}
    \label{tab:tab1}
\end{table}

\paragraph{Participants Satisfied with the Intuitive Design}
The results of study session three are presented in Table~\ref{tab:tab2}. Overall, participants expressed satisfaction with the current ``Intuitive'' design for different operations, with scores ranging from 4.8 to 7 across various criteria. They suggested that using a balance scale to represent comparison problems could further enhance engagement and reduce cognitive load. Additionally, they recommended including less text in the visuals to minimize cognitive load for learners. Our final design incorporates these suggestions, as detailed in Section~\ref{sec:design_element}.
\begin{table}[h]
    \centering
    \small
    \begin{tabular}{p{1.8cm}ccc}
     \toprule
      \textbf{Operation}   & \textbf{Clarity} & \textbf{Engagement} & \textbf{Cog Load Opt}\\
      \midrule
      Addition & 5.4  & 6.4 & 5.0\\
      Subtraction & 5.0  & 6.4 & 5.2\\
      Multiplication & \textbf{7.0}  & 6.4 & \textbf{6.0}\\
      Division & 6.4  & \textbf{6.6} & 5.4\\
      Surplus & 6.8  & \textbf{6.6} & 5.8\\
      Comparison & 5.6  & 6.0 & 4.8\\
      UnitTrans & 6.6  & \textbf{6.6} & 5.6\\
      MultiSteps & 6.6  & \textbf{6.6} & 5.8\\
    \midrule
    \end{tabular}
    \caption{Results of exploratory study session 3. They reflect experts' evaluations of the ``Intuitive'' design for different operations. All scores are on a 7-point Likert scale, where higher values indicate better performance.}
    \label{tab:tab2}
\end{table}

\paragraph{Potential Application of Our Visuals}
Participants suggested several potential applications for our visuals. They noted that our visuals can be easily attached to slides or textbooks and help with the following:

\begin{itemize} 
    \item \textbf{Facilitating MWP Understanding:} Four teachers mentioned that displaying our visuals in class can help students, especially those with learning difficulties, better access MWPs and build confidence in solving them. 
    \item \textbf{Enhancing Student Engagement:} Two teachers suggested using the visuals interactively---pointing to different entities in visuals and asking students to link them to corresponding parts of the MWP---can enhance engagement and learning. 
    \item \textbf{Teaching Mathematical Operations:} All five teachers agreed that the Intuitive design aids in teaching operations by representing them intuitively, thereby making abstract operations concrete and easier to understand.
\end{itemize}

\subsection{Details of Entity Visualization Design}
If the \texttt{entity\_quantity} does not exceed ten, we visualize each entity individually. For quantities greater than ten, we represent a single entity accompanied by the quantity number overlaid on it. This approach aligns with common designs in popular educational visual datasets like Twinkl~\citep{twinkl}.

\subsection{Details of Operation Visualization Designs}\label{sec:detail_operation}
Operations define the relationships between different containers. In addition to basic arithmetic operations such as addition, subtraction, multiplication, and division, we incorporate additional operations including surplus, comparison, unit transformation, and multi-step calculations. These operations enable our approach to cover 94.4\% of Grade 1-3 MWPs in the ASDiv dataset~\citep{asdiv2020}.

We visualize these operations using two visual variations: ``Formal'' and ``Intuitive''. In the ``Formal'' variation, operations are represented using mathematical symbols such as ``+'', ``-'', ``×'', and ``÷'', accompanied by text. We show examples in Appendix~\ref{sec:example_formal}.

In the ``Intuitive'' variation, each operation is represented through a specific visual arrangement (see visual examples in Appendix~\ref{sec:example_intuitive}): 

\noindent{\bf Addition:} Containers involved in the addition are enclosed within a rectangle. A purple circle with a question mark is placed at the bottom-right corner of the rectangle.

\noindent{\bf Subtraction}: The minuend container is visualized first, with the subtracted items crossed out. A purple circle with a question mark is placed at the bottom-right corner of the rectangle.

\noindent{\bf Multiplication}: The multiplicand container is visualized repeatedly to indicate multiplication. All entities are enclosed within a larger rectangle, with a purple circle and a question mark added at the bottom-right corner, similar to addition. A special type of multiplication involves computing ``area''. For such problems, we visualize it as a single item with dimensions corresponding to the width and length described in the MWP.

\noindent{\bf Division}: The division operation is visualized as the post-division state, with multiple container rectangles representing groups enclosed within a larger rectangle. If the MWP asks for the quantity per group (e.g., "10 apples divided into 5 boxes, how many per box?"), a purple question mark circle is placed at the bottom-right of the last container. If it asks for the number of groups (e.g., "10 apples, 2 per box, how many boxes?"), the question mark is placed at the top-right of the larger rectangle.

\noindent{\bf Surplus}: Similar to division, but the surplus container is visualized separately as the remainder. The remainder is placed at last, with a purple circle and a question mark at the bottom-right corner of its rectangle. 

\noindent{\bf Comparison}: This operation involves comparing different entities by visualizing them on a balance scale. Each container is placed on one side of the scale.

\noindent{\bf Unit Transformation}: We adopt a purple bubble above each visual item to display its value in the transformed unit.

Finally, for MWPs with multiple operations, we follow these visualization rules for each operation and dynamically combine them to form the overall expression tree (see Figure~\ref{fig:fig22}).
\section{Annotated Dataset Statistics} \label{sec:dataset}
We present the annotated dataset statistics in Table~\ref{tab:tab10}.

\begin{table*}
\small
\begin{tabular}{p{2cm}|M{1.5cm}|M{3.5cm}|M{4cm}|M{2cm}}
\hline
\textbf{Dataset} & \textbf{Visuals} & \textbf{Domain} & \textbf{Use Cases} & \textbf{Grade Level} \\
\hline
\sys{} (ours) & 1,903  & Primary School Math Word Problems & 
Supporting primary school students' math understanding; Evaluating and training Text-to-Image models on pedagogical visual generation & Primary school Grade 1-3 \\
\hline
MATH-Vision~\citep{wang2024measuring} & 3,040  & General mathematics, competition-level problems & 
Visual math problem-solving; Evaluating multimodal models on math reasoning & Middle school to high school (competition-level difficulty) \\
\hline
MathVista~\citep{lu2024mathvista} & 6,141  & Logical, algebraic, and scientific reasoning & 
Math visual question answering; Puzzle-solving ; logical reasoning; Function analysis ; diagram understanding & Varied (elementary to advanced reasoning) \\
\hline
\end{tabular}

\caption{Dataset Statistics}
\label{tab:tab10}
\label{tab:dataset_comparison}
\end{table*}

\section{Details of Rendering Programs}
We present the algorithm for rendering programs in Algorithm~\ref{alg:vl_to_svg}. We use rendering programs to map from \ac{VL} to the desirable visual. 
The rendering program first converts the \ac{VL} into a tree structure \(T\), where each operation becomes a parent node and each container becomes a child node. Next, we traverse \(T\) in a bottom-up manner. During this traversal, when a container node is encountered, its relative position is computed based on its attributes (e.g. the quantity of entity in this container). Conversely, when an operation node is encountered, the relative positions of its child nodes are updated according to the operation type. Note that the positioning of “Formal” and “Intuitive” visuals differs, as detailed in Section~\ref{sec:operation}. Once all relative positions are determined, a global layout plan is computed from these values. Finally, we traverse the tree in a top-down order and render each container and operation node according to the global layout plan, using the corresponding elements from the SVG dataset. We retrieve the SVG icon corresponding to the \texttt{entity\_type}, \texttt{container\_type} and \texttt{attr\_type} and map it as the source to the visual. The complete algorithm is presented in Algorithm~\ref{alg:vl_to_svg}.

\begin{algorithm}
    \caption{Rendering Visuals from MWP Visual Language}
    \label{alg:vl_to_svg}
    \begin{algorithmic}[1]
        \Require 
            \Statex $VL$: A visual language representation of the MWP
            \Statex $SVG$: A dataset of SVG elements for rendering
        \Ensure Rendered MWP visualization
        
        \State \textbf{Step 1: Parse VL} 
        \State Convert the visual language ($VL$) into a tree structure $T$, ignoring \texttt{result\_container} when generating ``Formal'' Visuals.
        
        \State \textbf{Step 2: Plan Layout}
        \For{each node $n$ in $T$ (traverse in bottom-up order)}
            \If{$n$ represents a container}
                \State Determine the relative position of $n$ based on its attributes (e.g., \texttt{entity\_type}, \texttt{entity\_quantity})
            \ElsIf{$n$ represents an operation}
                \State Update the relative position of $n$'s child node based on the operation type
            \EndIf
        \EndFor
        
        \State \textbf{Step 3: Compute Global Layout}
        \State Integrate the relative positions from all nodes to form a coherent global layout plan
        
        \State \textbf{Step 4: Render SVG}
        \For{each node $n$ in $T$ (traverse in top-down order)}
            \State Retrieve the final coordinates for $n$ from the global layout plan
            \State Render $n$ using the corresponding SVG element from the $SVG$ dataset
        \EndFor
        
    \end{algorithmic}
\end{algorithm}

\section{Generation Prompts}
\subsection{Prompt For Visual Language generation}\label{sec:prompt_vl}
We present the prompt we used for generating Visual Language from MWP as below:
\begin{lstlisting}
You are an expert in converting math word problems into a structured 'visual language'. Your task is to generate a visual language expression based on the given math word problem.

**Background Information**  
You should use the following fixed format for each problem:
<operation>(
    container1[entity_name: <name>, entity_type: <type>, entity_quantity: <number>, container_name: <container>, container_type: <container type>, attr_name: <attr>, attr_type: <attr type>],
    container2[entity_name: <name>, entity_type: <type>, entity_quantity: <number>, container_name: <container>, container_type: <container type>, attr_name: <attr>, attr_type: <attr type>],
    result_container[entity_name: <name>, entity_type: <type>, entity_quantity: <number>, container_name: <container>, container_type: <container type>, attr_name: <attr>, attr_type: <attr type>]
)

operation can be "addition", "subtraction", "multiplication", "division", "surplus", "area", "comparison", or "unittrans".

Each container has the attributes: entity_name, entity_type, entity_quantity, container_name, container_type, attr_name, attr_type.  
For example, a girl named Lucy may be represented as:  
entity_name: Lucy, entity_type: girl.  

The optional attributes container_name, container_type, attr_name, and attr_type allow extended descriptions.  
In the MWP description "Jake picked up three apples in the morning...", the container1 could be:  
entity_name: apple, entity_type: apple, entity_quantity: 3, container_name: Jake, container_type: boy, attr_name: morning, attr_type: morning.  
These additional attributes are not fixed and may vary according to different interpretations.

Example of Visual Languages: ...

Once you are ready to perform the task, you may write down your thought process, but please ensure that you provide the final visual language expression in the following format at the end:

visual_language: <the visual language result>  
Question: 
Solution expression:
\end{lstlisting}

\subsection{Prompt for Visual Generation}\label{sec:prompt_visual}
\subsubsection{Prompt for Formal Visual Generation}
\begin{lstlisting}
Please Create an educational visual for this math word problem: ...
Suppose this problem has solution expression: ...

The visual consists of:
1. Container: We use rectangular sections to represent different containers or group of entities. Inside each rectangle, display the entities of this container (e.g., apples, balls, etc.).
2. Container Name: Above each rectangle, place a container icon (e.g., an orange basket, jar, or other container type) and label it with the container's name (e.g., 'basket,' 'jar, etc).
3. Operation Symbol: Between each two rectangles, include an operation symbol that varies depending on the problem 
4. Outcome Section: To the right, place an '=' symbol followed by a '?' to symbolize the unknown solution.

Example:
For problem: Lucy has five oranges and Jake has two oranges. How many oranges do they have together?
Solution expression: 5+2=7
The visual consists of two containers, "Lucy" and "Jake," as rectangulars labeled with their names and icons (boy icon for Jake and girl for Lucy) on the top of each rectangle. Each rectangle contains oranges corresponding to their quantities (Lucy: 5, Jake: 2). A "+" symbol between the rectangles indicates the addition operation, and an "=" followed by a question mark represents the unknown solution.

Special cases:
1. For comparison problem, please use a balance scale to weigh different entities. For problem 'Lucy has 4 strawberries. Jake gave her 5 more. She needs 10 strawberries to make a cake. Does she have enough to make a cake?' We draw a balance scale. On the left side of the scale, two rectangular sections represent 'Lucy' and 'Jake,' each labeled with their names and icons. Lucy's section contains 4 strawberries, and Jake's section contains 5 strawberries. A "+" symbol indicates the addition of their strawberries. To the right of this, an "=" symbol and a question mark. On the right side of the scale, another rectangular section labeled "cake" contains 10 strawberries, representing the required amount. An "=" symbol and a question mark follow it. 
2. For unit transformation problem, please use a purple buble with the converted value in it on the top of each item to represent the unit value of the current item. For example, a problem like 'Charles found 6 pennies on his way to school. He also had 3 nickels already at home. How much money does he now have in all?' can be represented as a visual: on the left side, a rectangular section labeled "on his way" contains 6 pennies, each with a purple bubble above it displaying its converted value of 0.01 (representing dollars). On the right side, another rectangular section labeled "home" contains 3 nickels, each with a purple bubble above it displaying its converted value of 0.05. A "+" symbol is placed between the two sections to indicate the addition of their values. To the right of the sections, an "=" symbol is followed by a question mark.
3. For surplus problem, please use text remainder with a new question mark after previous question mark.
4. If any container have item quantity higher than 10, please visualize only one item inside this container rectangle to be bigger and put the quantity number to cover the item. For example, if the problem is 'Lucy has 15 apples and Jake has 3 apples. How many apples do they have together?', the visual should show 15 apples for Lucy and 3 apples for Jake. Lucy's apples should be represented by a single apple that is larger than Jake's apples, and the number 15 should be placed on top of it to indicate the quantity. Jake's apples should be represented by three smaller apples. The "+" symbol between the two entities indicates the addition operation, and an "=" symbol followed by a question mark.
\end{lstlisting}

\subsubsection{Prompt for Intuitive Visual Generation}
\begin{lstlisting}
    Please Create an educational visual for this math word problem: ...
Suppose this problem has solution expression: ...

The visual consists of:
1. Container: We use rectangular sections to represent different containers or group of items. Inside each rectangle, display the items of this container (e.g., apples, balls, etc.).
2. Container Name: Above each rectangle, place a container icon (e.g., an orange basket, jar, or other container type) and label it with the container's name (e.g., 'basket,' 'jar, etc).

Handle different operations:
1. For addition, use a big rectangle to cover all container rectangles need to be added together. And place a purple circle with question mark inside at the right bottom side of the big rectangle.
2. For subtraction, first visualize minuend container then cross out item that has been subtracted. Place a purple circle with question mark inside at the right bottom side of the minuend container rectangle.
3. For multiplication, repeatedly visualize the multiplicand container. Use a big rectangle to cover all container. Place purple circle with question mark similar as addition.
4. For division, visualize it as the state after division, with many container rectangles represent different groups. If asking about quantity in single container, place purple circle at the right bottom of the last container rectangle. If asking about number of container, place purple circle at the right top of the big rectangle.
5. For surplus, similar as division, only difference is you should visualize the surplus container at the last and place the purple circle at the right bottom side of surplus container rectangle.
6. For comparison, use a balance scale to weigh different containers. Visualize entities on the left and right side of the scale separately.
7. For unit transformation, use a purple buble with the converted value in it on the top of each item to represent the unit value of the current item.
8. For problem involving multiple addition and subtraction, use the same visualization rule and combine dynamically.

Example:
For problem: Lucy has five oranges and Jake has two oranges. How many oranges do they have together?
Solution expression: 5+2=7
The visual consists of two containers, "Lucy" and "Jake," as rectangulars labeled with their names and icons (boy icon for Jake and girl for Lucy) on the top of each rectangle. Each rectangle contains oranges corresponding to their quantities (Lucy: 5, Jake: 2). A bigger rectangle encompasses the two containers to indicate addition.
\end{lstlisting}

\begin{table*}
\centering
\scriptsize
\begin{tabular}{p{4.3cm}|M{1cm}M{1cm}|M{1cm}M{1cm}|M{1cm}M{1cm}|M{1cm}M{1cm}}
\hline
\textbf{Method} & \multicolumn{2}{c}{\textbf{Accuracy}} & \multicolumn{2}{c}{\textbf{Completeness}} & \multicolumn{2}{c}{\textbf{Clarity}} & \multicolumn{2}{c}{\textbf{Cog Load Opt}} \\
\hline
 & \textbf{Formal} & \textbf{Intuitive} & \textbf{Formal} & \textbf{Intuitive} & \textbf{Formal} & \textbf{Intuitive} & \textbf{Formal} & \textbf{Intuitive} \\
\hline
ft\_ mistral-7B-v0.3(E) & 2.83 & 2.71 & 3.00 & 2.67 & \textbf{3.00} & 2.71 & \textbf{2.96} & 2.71 \\
ft\_ mistral-7B-v0.3  & 2.54 & 2.08 & 2.67 & 2.04 & 2.67  & 2.08 & 2.63 & 2.08 \\
zs\_ mistral-7B-v0.3(E) & 1.33 & 1.00 & 1.38 & 1.00 & 1.46  & 1.00 & 1.46 & 1.00 \\
zs\_ mistral-7B-v0.3  & 1.25 & 1.00 & 1.21 & 1.00 & 1.29  & 1.00 & 1.33 & 1.00 \\
ft\_stable-diffusion-3.5-large(E)  & \textbf{2.96} & \textbf{2.88} & \textbf{3.12} & \textbf{3.08} & 2.92 & \textbf{3.75} & 2.83 & \textbf{3.58} \\
ft\_stable-diffusion-3.5-large & \textbf{2.96} & 2.75 & 3.08 & \textbf{3.08} & 2.79 & 3.58 & 2.83 & 3.54 \\
zs\_stable-diffusion-3.5-large(E)  & 2.71 & 2.67 & 2.96 & 2.92 & 2.83 & 3.08 & 2.83 & 2.96 \\
zs\_stable-diffusion-3.5-large & 2.71 & 2.67 & 2.96 & 2.71 & 2.83 & 3.08 & 2.83 & 2.96 \\
\hline
\end{tabular}
\caption{Other evaluation results for different visual generation methods. For each method, 48 visuals (24 Formal and 24 Intuitive) were evaluated, with each score representing the average rating from two researchers on a 1–5 scale (higher is better).  (E) indicates the method used the solution expression as input. ``ft'' means this model is fine-tuned on our annotated dataset, while ``zs'' means zero-shot.}
\label{tab:tab8}
\end{table*}
\section{Fine-tuning Details}\label{sec:fine-tune_detail}
\subsection{Fine-tuning LLMs}


To fine-tune LLMs for our task, we constructed a training set of 1,011 \ac{VL} instances using stratified sampling based on ``Grade'' and ``Question Type''. This represents 80\% of the full dataset. We fine-tuned four model variants: Llama-3.1-8B with and without solution expression, and Mistral-7B-v0.3 with and without solution expression.

We adopt the fine-tuning setup from \citet{wang-etal-2024-book2dial}, with modifications guided by prior work on parameter-efficient fine-tuning~\citep{Ding2023}. Each model is fine-tuned for 10 epochs with a per-device batch size of 2. We enable gradient checkpointing to reduce memory consumption and apply the paged\_adamw\_8bit optimizer~\citep{loshchilov2019decoupled} for efficient training. The learning rate is set to 2.5e-5, selected based on prior studies and pilot experiments that showed stable convergence. A linear decay scheduler is used with 3\% warmup steps to stabilize early training dynamics.

To ensure parameter efficiency, we incorporate LoRA adapters~\citep{hu2022lora}, which significantly reduce the number of trainable parameters while maintaining performance. All models are fine-tuned on a single NVIDIA RTX 4090 GPU. The Llama-3.1-8B models (with and without solution expression) contain approximately 8 billion parameters and require around 12 hours to train, while the Mistral-7B-v0.3 models require around 11 hours. For evaluation, we report the results of a human assessment of a single inference from each model.

\subsection{Fine-tuning Text-to-Image Models}
For fine-tuning TTI models, we create a Formal visual training set containing 1,011 visuals corresponding to the training set used by the LLM, and an Intuitive visual training set containing 502 visuals. Both training sets occupy 80\% of their corresponding ground truth datasets. We fine-tuned Flux.1-dev and Stable Diffusion-3.5-large, both with and without solution expression---resulting in four TTI model variants in total.
Each model was fine-tuned for 10 epochs using a batch size of 5. This batch size was selected to balance GPU memory constraints (on a single RTX 4090) and batch diversity, which we found to improve convergence stability in preliminary runs. We enabled gradient checkpointing to reduce memory consumption and allow deeper model tuning without sacrificing input resolution.

We used the AdamW\_BF16 optimizer~\citep{loshchilov2019decoupled} with an initial learning rate of 1e-5, a commonly effective starting point in vision-language fine-tuning~\citep{yeh2023navigating}, particularly for stable convergence when using BF16 precision. Learning rate was decayed using a polynomial scheduler with no warmup; this choice was empirically motivated by smoother training dynamics observed in ablation runs, compared to linear decay or cosine schedules.

To enable parameter-efficient fine-tuning, we adopted LoRA adapters via the Lycoris framework~\citep{yeh2023navigating}, which allowed us to adapt attention layers without full weight updates—crucial given the size of the models (8.1B–12B parameters). Images were generated at 1024×1024 resolution, which aligns with the target use case of producing high-quality visuals for educational settings, while remaining computationally feasible.

All fine-tuning was performed on a single NVIDIA RTX 4090 GPU. The Flux.1-dev models (12B parameters) required around 48 hours to train, while the Stable Diffusion-3.5-large models (8.1B parameters) completed training in around 15 hours. For evaluation, we report results based on human judgments of single inference outputs per model.

\subsection{Other Fine-tuning Results}
We present other fine-tuning experiment results in Table~\ref{tab:tab8}.
\section{Details of Qualitative Analysis}
\subsection{Procedure}\label{sec:open_coding}
The thematic analysis was performed on a sample of 120 visuals generated by the fine-tuned Flux model with expression, zero-shot Flux model with expression and Recraft-v3 with expression, and a total of eight error types were identified. However, three of these error types occurred fewer than eight times. After discussing the findings, we consolidated the labels and focus on five major types of error in close coding.

In the systematic evaluation phase, two researchers manually analyzed 576 visuals generated by the fine-tuned Flux model with expression, the zero-shot Flux model with expression, and Recraft-v3 with expression. 

\subsection{Statistical Results}\label{sec:stat_results}
We present the statistical results for supporting qualitative analysis in Table~\ref{tab:tab11}. The results aggregated by Grade is shown in Table~\ref{tab:tab13} and results aggregated by operation is shown in Table~\ref{tab:tab14}.

\begin{table*}[t]
\centering
\scriptsize
\begin{tabular}{p{1.2cm}c
M{0.9cm}M{0.9cm}  
M{0.9cm}M{0.9cm}  
M{0.9cm}M{0.9cm}  
M{0.9cm}M{0.9cm}  
M{0.9cm}M{0.9cm}  
}
\toprule
\textbf{Method} & \textbf{Grade} 
& \multicolumn{2}{c}{\textbf{Quantity Err}} 
& \multicolumn{2}{c}{\textbf{Relation Err}} 
& \multicolumn{2}{c}{\textbf{Struct Misalign}} 
& \multicolumn{2}{c}{\textbf{Miss Visual Item}} 
& \multicolumn{2}{c}{\textbf{Miss Context Cue}} \\
\cmidrule(lr){3-4}
\cmidrule(lr){5-6}
\cmidrule(lr){7-8}
\cmidrule(lr){9-10}
\cmidrule(lr){11-12}
 & 
 & \textbf{Formal} & \textbf{Intuitive} 
 & \textbf{Formal} & \textbf{Intuitive} 
 & \textbf{Formal} & \textbf{Intuitive} 
 & \textbf{Formal} & \textbf{Intuitive} 
 & \textbf{Formal} & \textbf{Intuitive} \\
\midrule
\multirow{3}{*}{ft\_flux.1-dev(E)} 
& 1 & 0.77 & 0.67 & 0.92 & 0.81 & 0.46 & 0.33 & 0.39 & 0.29 & 0.77 & 0.38 \\
& 2 & 0.67 & 0.80 & 0.83 & 0.90 & 0.33 & 0.23 & 0.38 & 0.33 & 0.42 & 0.43 \\
& 3 & 0.73 & 0.73 & 0.85 & 0.76 & 0.34 & 0.18 & 0.48 & 0.29 & 0.59 & 0.58 \\
\midrule
\multirow{3}{*}{zs\_flux.1-dev(E)} 
& 1 & 0.85 & 0.76 & 0.85 & 0.86 & 1.00 & 1.00 & 0.39 & 0.52 & 0.85 & 0.48 \\
& 2 & 0.75 & 0.70 & 0.88 & 0.73 & 1.00 & 1.00 & 0.67 & 0.73 & 0.46 & 0.50 \\
& 3 & 0.76 & 0.84 & 0.95 & 0.93 & 1.00 & 1.00 & 0.85 & 0.67 & 0.64 & 0.73 \\
\midrule
\multirow{3}{*}{recraft-v3(E)} 
& 1 & 0.39 & 0.33 & 0.85 & 0.76 & 0.69 & 0.95 & 0.39 & 0.14 & 0.54 & 0.33 \\
& 2 & 0.33 & 0.43 & 0.79 & 0.80 & 0.50 & 0.93 & 0.42 & 0.23 & 0.25 & 0.40 \\
& 3 & 0.44 & 0.36 & 0.83 & 0.84 & 0.68 & 0.93 & 0.46 & 0.16 & 0.36 & 0.64 \\
\bottomrule
\end{tabular}
\caption{Statistical Results for Qualitative Analysis by Grade.}
\label{tab:tab13}
\vspace{-8pt}
\end{table*}
\begin{table*}[t]
\centering
\scriptsize
\begin{tabular}{p{1.2cm}p{1cm}
M{0.8cm}M{0.8cm}  
M{0.8cm}M{0.8cm}  
M{0.8cm}M{0.8cm}  
M{0.9cm}M{0.9cm}  
M{0.9cm}M{0.9cm}  
}
\toprule
\textbf{Method} & \textbf{Operation} 
& \multicolumn{2}{c}{\textbf{Quantity Err}} 
& \multicolumn{2}{c}{\textbf{Relation Err}} 
& \multicolumn{2}{c}{\textbf{Struct Misalign}} 
& \multicolumn{2}{c}{\textbf{Miss Visual Items}} 
& \multicolumn{2}{c}{\textbf{Miss Context Cues}} \\
\cmidrule(lr){3-4}
\cmidrule(lr){5-6}
\cmidrule(lr){7-8}
\cmidrule(lr){9-10}
\cmidrule(lr){11-12}
 & 
 & \textbf{Formal} & \textbf{Intuitive} 
 & \textbf{Formal} & \textbf{Intuitive} 
 & \textbf{Formal} & \textbf{Intuitive} 
 & \textbf{Formal} & \textbf{Intuitive} 
 & \textbf{Formal} & \textbf{Intuitive} \\
\midrule
\multirow{8}{*}{ft\_flux.1-dev(E)} 
& addition      & 0.67 & 0.59 & 0.67 & 0.70 & 0.28 & 0.25 & 0.33 & 0.25 & 0.61 & 0.41 \\
& comparison    & 1.00 & 0.80 & 0.80 & 1.00 & 0.40 & 0.00 & 0.60 & 0.00 & 0.40 & 0.40 \\
& division      & 0.60 & 0.90 & 0.90 & 0.90 & 0.40 & 0.20 & 0.60 & 0.30 & 0.50 & 0.70 \\
& multiplication& 0.54 & 0.67 & 0.69 & 0.67 & 0.31 & 0.17 & 0.23 & 0.58 & 0.08 & 0.50 \\
& subtraction   & 0.71 & 0.88 & 0.95 & 1.00 & 0.19 & 0.25 & 0.38 & 0.38 & 0.76 & 0.38 \\
& surplus       & 1.00 & 1.00 & 1.00 & 1.00 & 0.80 & 0.38 & 0.40 & 0.12 & 0.60 & 0.75 \\
& unittrans     & 1.00 & 1.00 & 1.00 & 1.00 & 0.75 & 0.00 & 0.75 & 0.67 & 0.00 & 0.33 \\
& multisteps    & 0.75 & 1.00 & 0.95 & 1.00 & 0.40 & 0.33 & 0.55 & 0.33 & 0.85 & 0.67 \\
\midrule
\multirow{8}{*}{zs\_flux.1-dev(E)} 
& addition      & 0.67 & 0.68 & 0.78 & 0.75 & 1.00 & 1.00 & 0.72 & 0.66 & 0.61 & 0.50 \\
& comparison    & 1.00 & 1.00 & 1.00 & 1.00 & 1.00 & 1.00 & 0.60 & 0.20 & 0.60 & 0.80 \\
& division      & 0.90 & 0.80 & 1.00 & 0.90 & 1.00 & 1.00 & 0.90 & 0.70 & 0.60 & 0.80 \\
& multiplication& 0.46 & 0.83 & 0.92 & 1.00 & 1.00 & 1.00 & 0.62 & 1.00 & 0.23 & 0.67 \\
& subtraction   & 0.81 & 0.63 & 0.90 & 0.75 & 1.00 & 1.00 & 0.67 & 0.62 & 0.81 & 0.62 \\
& surplus       & 1.00 & 1.00 & 1.00 & 1.00 & 1.00 & 1.00 & 0.80 & 0.25 & 0.60 & 0.75 \\
& unittrans     & 1.00 & 1.00 & 1.00 & 1.00 & 1.00 & 1.00 & 0.75 & 0.67 & 0.00 & 0.33 \\
& multisteps    & 0.80 & 1.00 & 0.95 & 1.00 & 1.00 & 1.00 & 0.85 & 0.83 & 0.85 & 0.67 \\
\midrule
\multirow{8}{*}{recraft-v3(E)} 
& addition      & 0.44 & 0.20 & 0.72 & 0.73 & 0.67 & 0.93 & 0.33 & 0.20 & 0.22 & 0.36 \\
& comparison    & 0.60 & 0.60 & 0.40 & 0.80 & 0.40 & 0.80 & 0.60 & 0.00 & 0.40 & 0.40 \\
& division      & 0.40 & 0.50 & 0.80 & 0.90 & 0.50 & 1.00 & 0.40 & 0.20 & 0.60 & 0.80 \\
& multiplication& 0.54 & 0.33 & 0.77 & 0.83 & 0.77 & 0.83 & 0.31 & 0.42 & 0.08 & 0.58 \\
& subtraction   & 0.24 & 0.25 & 0.90 & 0.88 & 0.62 & 1.00 & 0.33 & 0.00 & 0.57 & 0.50 \\
& surplus       & 0.60 & 0.63 & 0.60 & 0.88 & 0.40 & 1.00 & 0.00 & 0.00 & 0.40 & 0.75 \\
& unittrans     & 0.50 & 1.00 & 1.00 & 1.00 & 0.75 & 1.00 & 1.00 & 0.33 & 0.00 & 0.33 \\
& multisteps    & 0.35 & 0.83 & 1.00 & 1.00 & 0.70 & 1.00 & 0.70 & 0.00 & 0.35 & 0.67 \\
\bottomrule
\end{tabular}
\caption{Statistical Results for Qualitative Analysis by operation.}
\label{tab:tab14}
\vspace{-8pt}
\end{table*}


\begin{table}[h]
    \centering
    \small
    \begin{tabular}{p{3.2cm}p{1.5cm}p{1.7cm}}
     \toprule
      \textbf{Criterion}   & \textbf{Edit Dist$\downarrow$}  & \textbf{LM Ratio$\uparrow$} \\
      \midrule
    ft\_ mistral-7B-v0.3(E) & \textbf{2.98} & \textbf{39.30} \\
    ft\_ mistral-7B-v0.3 & 3.14 & 19.07 \\
    zs\_ mistral-7B-v0.3(E) & 7.05 & 0.00 \\
    zs\_ mistral-7B-v0.3 & 6.86 & 0.00 \\
    \hline
    \end{tabular}
    \caption{Other results of Visual Language generation. E denotes generation with the solution expression as input.}
    \label{tab:tab9}
\end{table}

\section{Ethical Consideration and Applications}
\subsection{Potential Risks}
One potential risk is that the generated visuals might be misinterpreted if they do not accurately capture the intended mathematical relationships, potentially leading to confusion among students and educators. To minimize this risk, we collaborated closely with primary school math teachers to develop the structured design space that aligns with pedagogical standards. We further annotate the generated dataset and ensure clarity and accuracy in the visuals.

\subsection{Terms of Use}

This section outlines the terms and conditions for the use of \sys{}. By using the code and datasets in this project, users agree to the following terms:

\paragraph{Prohibited Use}

The code and datasets shall not be used for commercial purposes without prior written consent from the authors.

\paragraph{Attribution}
When using or referencing the code and datasets, users must provide proper attribution to the original authors.

\paragraph{No Warranty}
This project is provided as is without any warranties of any kind, either expressed or implied, including but not limited to fitness for a particular purpose. The authors are not responsible for any damage or loss resulting from the use of this project.

\paragraph{Liability}
The authors shall not be held liable for any direct, indirect, incidental, special, exemplary, or consequential damages arising in any way out of the use of the \sys{} project.

\paragraph{Updates and Changes}
The authors reserve the right to make changes to the terms of this license or the \sys{} itself at any time.

\subsection{Compliance with Artifact Usage and Intended Use Specifications}

\subsubsection{Compliance with Existing Artifact Usage}
In our study, we utilized a range of existing artifacts, such as open-source SVG datasets from various sources~\citep{svgrepoRepoFree,iconfont2025,svgen2025, kaggleIcons, huggingfaceUmuthopeyildirimsvgen500kDatasets, pexels} and ASDiv dataset~\citep{asdiv2020}, to develop our visual datasets. We rigorously ensured that our usage of these materials was in strict accordance with their intended purposes, aligning with each dataset's vision of freely accessible content. Additionally, we employed various computational tools within their prescribed licensing terms, thus adhering to ethical and legal standards.

\subsubsection{Specification of Intended Use for Created Artifacts}
Our research led to the development of two significant artifacts:

\paragraph{Framework for Generating Pedagogically Meaningful Visuals}\mbox{} \par
\textbf{Intended Use:} This framework is designed for academic research and educational technology development. It facilitates the generation of pedagogically meaningful visuals, aiming to enhance AI-driven educational tools.

\textbf{Restrictions:} The framework should be used within the bounds of educational and research settings. Any commercial or high-stakes educational application is advised against without further validation and ethical review.

\textbf{Ethical Considerations:} We emphasize the responsible use of this framework, particularly in maintaining the integrity and context of the source textbooks.

\paragraph{Dataset of Generated Visuals}\mbox{} \par
\textbf{Intended Use:} The dataset is primarily intended for research in educational technologies. It offers a resource for developing and testing Text-to-Image models in educational contexts.

\textbf{Restrictions:} This dataset is not recommended for direct application in live educational settings without substantial vetting, as it may contain synthetic inaccuracies.

\textbf{Data Ethics:} As the dataset is derived from open-source SVG datasets, it respects the principles of open access. We encourage users to keep the dataset within academic and research domains, in line with the ethos of the source material.

\subsection{Data Collection and Anonymization Procedures}
In our research, rigorous steps were taken to ensure that the data collected and used did not contain any personally identifiable information or offensive content. The data, primarily sourced from open-access MWP datasets and SVG datasets, inherently lacked individual personal data. For the components involving human interaction, such as feedback or evaluation, all identifying information was carefully removed to maintain anonymity. Additionally, we implemented a thorough review process to screen for and exclude any potentially offensive or sensitive material from our dataset. These measures were taken to uphold the highest standards of privacy, ethical data usage, and respect for individual confidentiality.

\subsection{Artifact Documentation}

\subsubsection{Visual Generation Framework}

\paragraph{Domain Coverage} The framework is designed to generate pedagogically meaningful visuals from MWP for teaching MWP.
\paragraph{Operation Coverage} It covers seven operations including: addition, subtraction, multiplication, division, surplus, comparison and unit transformation.

\subsubsection{Dataset of Generated Visuals}

\paragraph{Visual and Style} The visuals are primarily generated from English MWPs. The style is educational and academic, suited for educational purposes.
\paragraph{Content Diversity} The dataset spans multiple academic disciplines, offering a rich variety of topics and themes.

\paragraph{Demographic Representation} While the dataset itself does not directly represent demographic groups (as it is synthesized from MWP dataset), the diversity in the source material reflects a broad spectrum of cultural and societal contexts.

\subsection{Use of AI Assistants in Research}
In our study, AI assistants were used sparingly and in accordance with ACL's Policy on AI Writing Assistance. We utilized ChatGPT and Grammarly for basic paraphrasing and grammar checks, respectively. These tools were applied minimally to ensure the authenticity of our work and to adhere strictly to the regulatory standards set by ACL. Our use of these AI tools was focused, responsible, and aimed at supplementing rather than replacing human input and expertise in our research process.

\subsection{Instructions Given To Participants}
\subsubsection{Disclaimer for Annotators}
Thank you for participating in our evaluation process. Please read the following important points before you begin:

\begin{itemize}
    \item \textbf{Voluntary Participation:} Your participation is completely voluntary. You have the freedom to withdraw from the task at any time without any consequences.
    \item \textbf{Confidentiality:} All data you will be working with is anonymized and does not contain any personal information. Your responses and scores will also be kept confidential.
    \item \textbf{Risk Disclaimer:} This task does not involve any significant risks. It primarily consists of reading and scoring generated visuals.
    \item \textbf{Queries:} If you have any questions or concerns during the task, please feel free to reach out to us.
\end{itemize}

\subsubsection{Instructions for Experiments}

Thank you for participating in our study. This research has received ethical approval, and your consent has been obtained. The entire study will take approximately 1.5 to 2 hours and consists of four sessions. Please read the instructions below carefully:

\paragraph{Session One – Visual Approach Preference:}
You will be shown two visual approaches for representing math word problems (MWPs):

1. Multiple Visuals: Each visual represents one sentence of the MWP.

2. Single Visual: One visual represents the entire MWP.
Please indicate your preference between these two approaches.

\paragraph{Session Two – Design Variation Evaluation:}
You will review six design variations for visualizing MWPs. These variations differ based on:

\begin{enumerate}
\itemsep0em 
    \item How Quantities Are Visualized:
    \begin{itemize}
        \item \textbf{Abstract}: Quantities are represented as text from the MWP.
        \item \textbf{Hybrid}: A single item is visualized with a label at the bottom-right corner indicating its quantity.
        \item \textbf{Visual}: Items are directly drawn in quantities matching their number.
    \end{itemize}
    \item How Operations Are Visualized:
    \begin{itemize}
        \item \textbf{Formal}: Mathematical operations are represented using standard symbols (e.g., +, -, ×, ÷).
        \item \textbf{Intuitive}: Operations are visualized using specific arrangements for each operation.
    \end{itemize}
\end{enumerate}
For each design variation, please complete a questionnaire rating:

\begin{itemize}
    \item Clarity: How clearly the visual design represents the MWP.
    \item Engagement: Whether the design appears to improve student engagement.
    \item Cognitive Load: Whether the design avoids introducing unnecessary cognitive load.
\end{itemize}
The order of presentation will be randomized to minimize order effects.

\paragraph{Session Three – Operation Design Feedback:}
In this session, you will review our ``Intuitive'' design for visualizing mathematical operations. Using the same criteria (Clarity, Engagement, Cognitive Load), please provide your feedback via a questionnaire. The presentation order will also be randomized.

\paragraph{Session Four – Evaluation Criteria Discussion:}
We will discuss with you the criteria that will be used to analyze the visuals generated by our system. This discussion focuses on how effectively our automated generation approach reproduces the intended design. After this discussion, you will complete a post-task questionnaire assessing the pedagogical value of the visual design.

Please answer all questions honestly and provide any suggestions for improvement. Your feedback is crucial for enhancing our framework. If you have any questions during the study, feel free to ask the researcher.

Thank you for your time and valuable input!

\subsubsection{Data Consent}
The data you provide during this study will be used solely for academic research purposes. All information will be anonymized and securely stored, and any published or shared data will be aggregated to ensure your privacy. By participating, you agree to the use of your data as described, but you retain the right to withdraw your consent at any time without penalty. If you have any questions about how your data will be used, please feel free to ask the research team.

\end{document}